%% file: main.tex
\PassOptionsToPackage{svgnames}{xcolor}
\documentclass[10pt,twocolumn,letterpaper]{article}
\usepackage{times}
\input{math_commands.tex}

\usepackage{xurl}
\usepackage[numbers]{natbib}
\PassOptionsToPackage{numbers, sort&compress}{natbib}
\usepackage{graphicx}
\usepackage{svg}
\usepackage[utf8]{inputenc} 
\usepackage[T1]{fontenc}    
\usepackage{booktabs}       
\usepackage{amsfonts,amssymb}       
\usepackage{nicefrac}       
\usepackage{microtype}      
\usepackage{graphicx}
\usepackage{booktabs}
\usepackage{makecell}
\usepackage{multirow}
\usepackage{enumitem}
\usepackage{floatflt}
\usepackage{adjustbox}
\usepackage[rightcaption]{sidecap}
\usepackage{array}
\usepackage{makecell}
\usepackage{comment}
\usepackage{placeins}
\usepackage{bbding}
\usepackage{relsize}
\usepackage{listings}
\usepackage{subcaption}
\usepackage[pagebackref,breaklinks,colorlinks]{hyperref}

\definecolor{VeryLightGray}{RGB}{240,240,240}
\definecolor{ReviseGreen}{RGB}{34, 139, 76}
\lstdefinestyle{PythonStyle}{
    language=Python,
    basicstyle=\footnotesize\ttfamily,
    keywordstyle=\color{Green},
    stringstyle=\color{Red},
    commentstyle=\color{Blue},
    backgroundcolor=\color{VeryLightGray},
    frame=none,
    numbers=left,
    numberstyle=\tiny\color{gray},
    numbersep=3pt,
}
\lstdefinestyle{BashStyle}{
    language=Bash,
    basicstyle=\footnotesize\ttfamily,
    keywordstyle=\color{Blue},
    stringstyle=\color{Red},
    commentstyle=\color{Gray},
    backgroundcolor=\color{VeryLightGray},
    frame=none,
    numbers=left,
    numberstyle=\tiny\color{gray},
    numbersep=3pt,
    emphstyle=\color{Purple},
    emph={python},
}

\makeatletter
\renewcommand{\@fnsymbol}[1]{}
\makeatother
\newboolean{blind}
\setboolean{blind}{false}
\usepackage{amsmath}
\usepackage[capitalize]{cleveref} 
\usepackage{tabularx}
\title{\ifthenelse{\boolean{blind}}{}{\textit{NeuCo-Bench}: }A Novel Benchmark Framework for Neural Embeddings in Earth Observation}

\author{%
  Rikard Vinge$^{1\ast}$, Isabelle Wittmann$^{5\ast}$, Jannik Schneider$^3$,\\
  Michael Marszalek$^1$, Luis Gilch$^4$, Thomas Brunschwiler$^5$, Conrad M Albrecht$^{1,2}$\\[1ex]
  $^1$ German Aerospace Center, Germany\\
  $^2$ Columbia University, USA\\
  $^3$ Juelich Supercomputing Center, Germany\\
  $^4$ IBM, Germany\\
  $^5$ IBM Research -- Europe, Switzerland
\thanks{\vspace{1.5ex}\relsize{-.5}
$^\ast$: equal contribution, CRediT \url{https://credit.niso.org}:\newline
\textbf{Rikard Vinge}: Methodology, Data curation, Software, Investigation, Validation, Visualization, Writing–-draft, Writing–-review \& editing;
\textbf{Isabelle Wittmann}: Methodology, Software, Investigation, Validation, Visualization, Writing-–draft, Writing-–review \& editing;
\textbf{Jannik Schneider}: Software, Investigation, Validation, Visualization, Writing-–review \& editing;
\textbf{Michael Marszalek}: Data curation, Software, Investigation, Validation, Visualization, Writing–-review \& editing;
\textbf{Luis Gilch}: Software, Validation;
\textbf{Thomas Brunschwiler}: Resources, Writing--review \& editing, Supervision, Funding acquisition, Project administration;
\textbf{Conrad M Albrecht}: Conceptualization, Methodology, Formal Analysis, Software, Data curation, Visualization, Resources, Writing--draft, Writing--review \& editing, Supervision, Funding acquisition, Project administration
}
}

\usepackage{enumitem}
\usepackage{float}

\begin{document}
\setlength{\parskip}{3pt}
\setlist{nosep}

\date{}
\maketitle

\begin{abstract}
We introduce \ifthenelse{\boolean{blind}}{}{\textit{NeuCo-Bench}, }a novel benchmark framework for evaluating (lossy) neural compression and representation learning in the context of Earth Observation (EO). Our approach builds on fixed-size embeddings that act as compact, task-agnostic representations applicable to a broad range of downstream tasks. \ifthenelse{\boolean{blind}}{Our benchmark}{NeuCo-Bench} comprises three components: (i) an evaluation pipeline built around embeddings, (ii) a challenge mode with a hidden-task leaderboard designed to mitigate pretraining bias, and (iii) a scoring system that balances accuracy and stability. To support reproducibility, we release \ifthenelse{\boolean{blind}}{}{SSL4EO-S12-downstream, }a curated multispectral, multitemporal EO dataset. We present results from a public challenge at \ifthenelse{\boolean{blind}}{a}{the 2025 CVPR EarthVision} workshop and conduct ablations with state-of-the-art foundation models. \ifthenelse{\boolean{blind}}{Our benchmark}{NeuCo-Bench} provides a step towards community-driven, standardized evaluation of neural embeddings for EO and beyond.
\vspace{-2ex}
\end{abstract}

\section{Introduction}
\input{sects/introduction}

\section{Related Work}
\input{sects/related_work}

\begin{figure*}[!ht]
  \centering
  \ifthenelse{\boolean{blind}}
  {\includegraphics[width=0.95\textwidth]{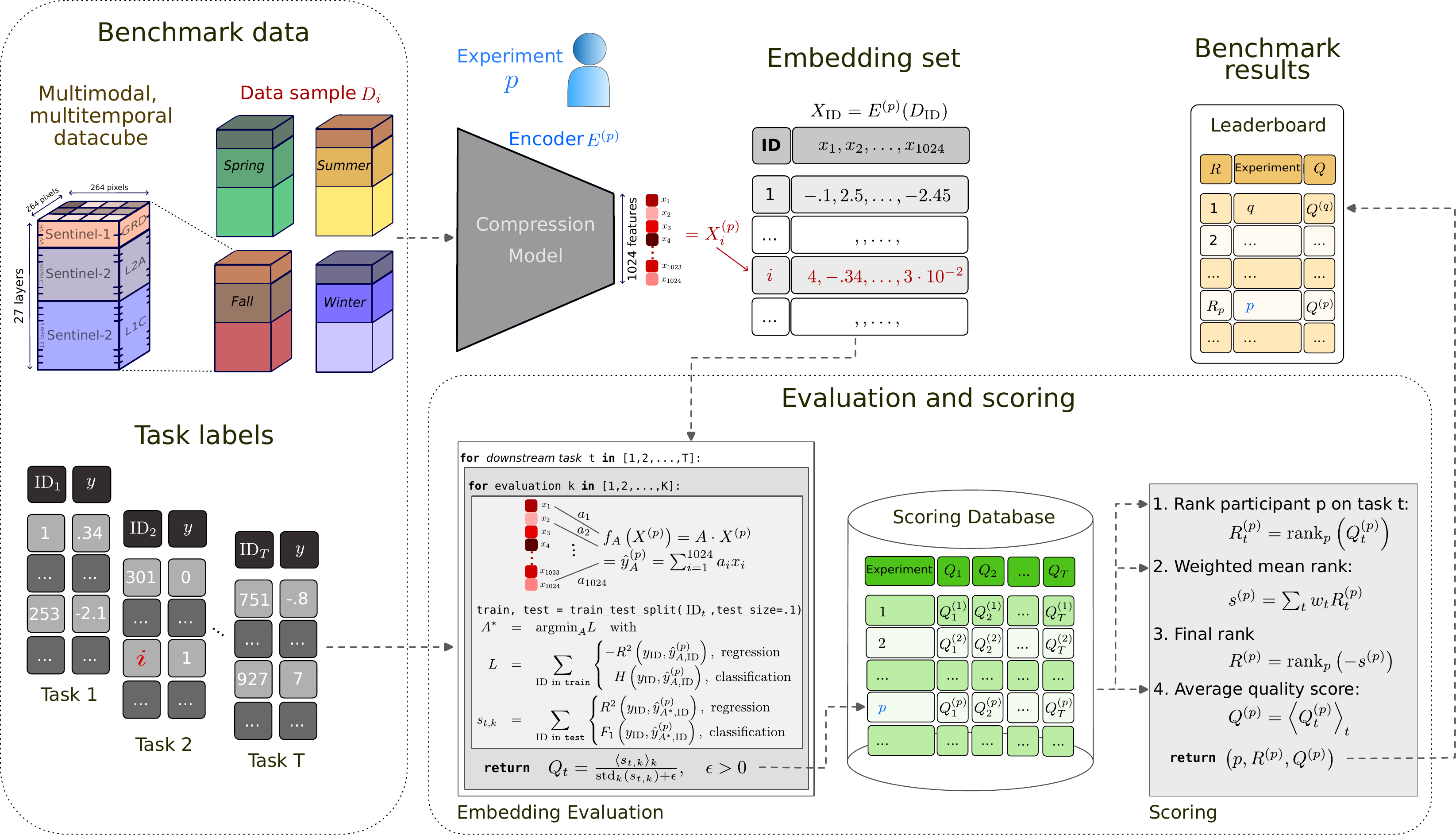}}
  {\includegraphics[width=0.95\textwidth]{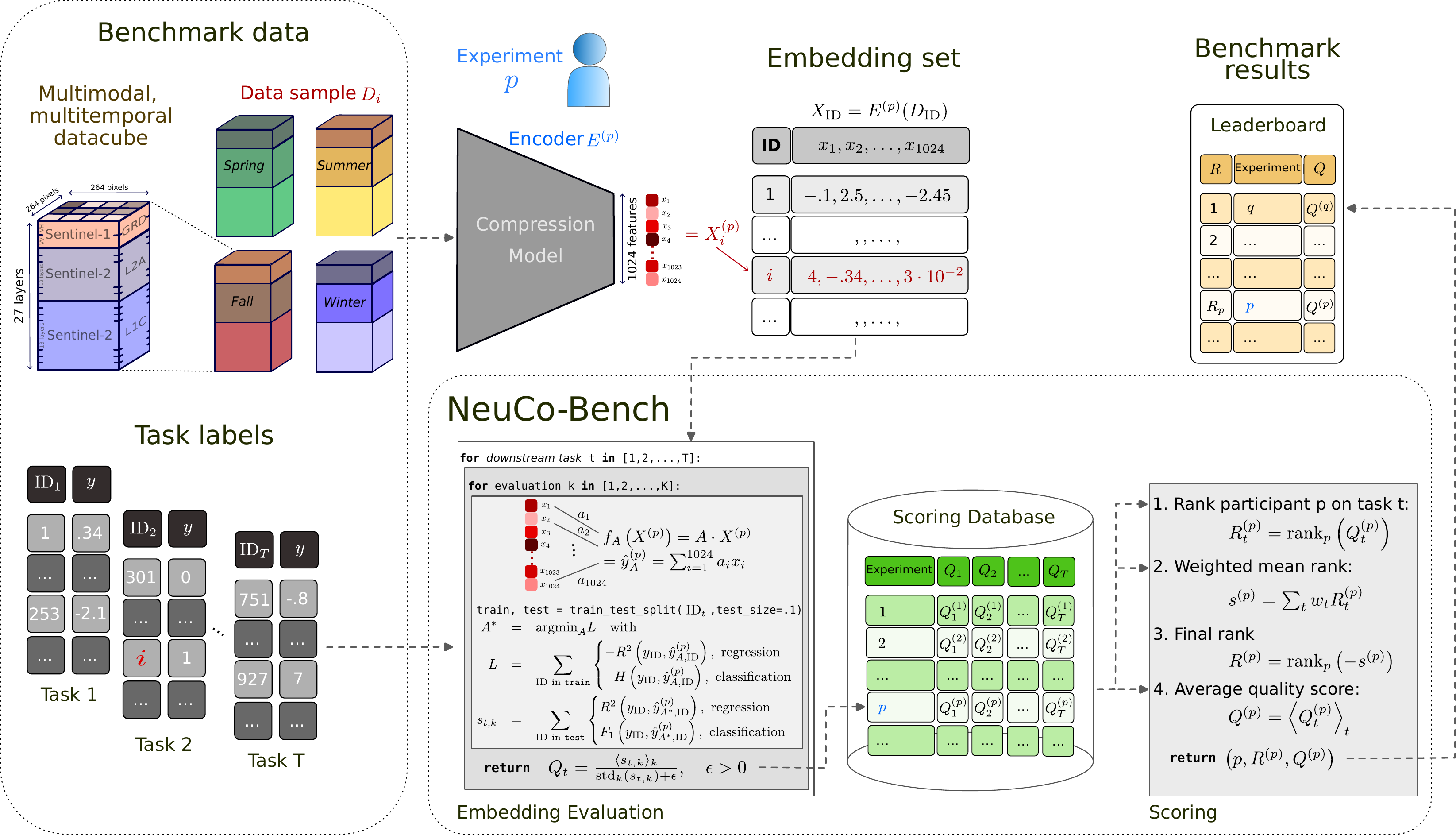}}
  \caption{Workflow diagram: The user of our benchmark compresses a set of downstream data into fixed-length embeddings (size $N=1024$ here). Our benchmark loads the embeddings for each of the $T$ downstream tasks, performs $K$ evaluations, randomly sampling a training and test split from the task data each evaluation, and scores the result compared to all previous experiments as stored in the \textit{Scoring Database}.}
  \label{fig_evaluation_scheme}
  \vspace{-2ex}
\end{figure*}

\section{Benchmarking Framework}
\label{sec:BenchmarkFramework}
\input{sects/framework}

\section{Benchmark Tasks}
\label{sec:BenchmarkTasks}
\input{sects/downstream_data}

\section{Benchmark Evaluation}
\label{sec:evaluation}

We evaluate the benchmark in two parts. Firstly, Section~\ref{sec:exp_setup} details the real-world evaluation of the benchmark in \ifthenelse{\boolean{blind}}{a}{the 2025 CVPR EarthVision} data challenge, as well as discussion of outcomes and learnings in Section~\ref{sec:eval_cvpr}. Secondly, Section~\ref{sec:eval_general} presents baseline evaluations exploring our set of downstream tasks.

\input{sects/evaluations_cvpr}
\input{sects/evaluations_general}

\input{sects/limitations_futurework}

\section{Conclusion}
We present \ifthenelse{\boolean{blind}}{}{NeuCo-Bench,} a task-driven benchmarking framework for evaluating compact embeddings by downstream task performance.
The framework introduces a novel rank-then-aggregate scoring method which dynamically determines the task complexity based on score statistics. We demonstrated \ifthenelse{\boolean{blind}}{our}{the NeuCo-Bench} framework by running \ifthenelse{\boolean{blind}}{a}{the 2025 CVPR EarthVision} data challenge on multi-modal, multi-temporal, and open-source EO data and introducing a set of real-world downstream tasks which remained undisclosed at the time of the data challenge, and have been publicly released after the conclusion of the competition. \ifthenelse{\boolean{blind}}{Our benchmark}{NeuCo-Bench} encourages the development of methods that generate semantically-rich, general-purpose embeddings.

We demonstrate that modern EO FMs, such as the multi-modal model TerraMind yield strong overall performance---particularly on semantic land-cover tasks. Post-encoding fusion of seasonal views results in notable gains for temporally sensitive tasks such as cloud cover prediction. We also observe that smaller and, in some cases, larger embedding sizes may degrade performance. This observation highlights compact embeddings as a practical choice for image-level tasks when high-quality annotations and compute resource become scarce.

\ifthenelse{\boolean{blind}}{Our benchmark}{NeuCo-Bench} is open source and ready for extension---either by novel evaluation methods or additional downstream tasks without any conceptual restriction to Earth observation. Currently, \ifthenelse{\boolean{blind}}{our framework}{NeuCo-Bench} is limited to image-level tasks, but future work aims to extend the functionality to include pixel-wise outputs, options beyond linear probing, and an assessments of bit-rate efficiency.

\ifthenelse{\boolean{blind}}{}{
\section*{Acknowledgment}
This work was funded by \textit{Embed2Scale} which is co-funded by the EU Horizon Europe program under Grant Agreement No.~\texttt{101131841}. Additional funding for this project has been provided by the Swiss State Secretariat for Education, Research and Innovation (SERI) and UK Research and Innovation (UKRI).

We thank Tim Reichelt (Oxford U) and Damien Robert (U Zurich) for support in creating labels for the downstream tasks. Johannes Jakubik and Benedikt Blumenstiel (IBM Research) provided valuable discussions when writing this article. CMA is grateful for Columbia University's academic hospitality in 2025.

Moreover, we are grateful for the active participation in our 2025 CVPR EarthVision workshop data challenge with valuable Q\&A in \url{https://github.com/DLR-MF-DAS/embed2scale-challenge-supplement/issues?q=is\%3Aissue}. In particular, conversations with Burak Ekim (UniBW) and Isaac Corley (Wherobots) sparked lively discussions, and our meetings with the winning teams \texttt{KTH and Friends} (\textit{KTH Royal Institute of Technology}, Sweden) and \texttt{404 Embedding Not Found} (Microsoft Research, USA).
}
\clearpage
\bibliographystyle{plainnat}

\input{main.bbl}
\clearpage
\appendix
\input{supplement-contents}

\end{document}

%% file: math_commands.tex
\usepackage{amsmath,amsfonts,bm}

\def\eqref#1{equation~\ref{#1}}

\def\1{\bm{1}}

\DeclareMathAlphabet{\mathsfit}{\encodingdefault}{\sfdefault}{m}{sl}
\SetMathAlphabet{\mathsfit}{bold}{\encodingdefault}{\sfdefault}{bx}{n}



%% file: sects/introduction.tex
\begin{figure*}[t]
    \includegraphics[width=0.98\textwidth]{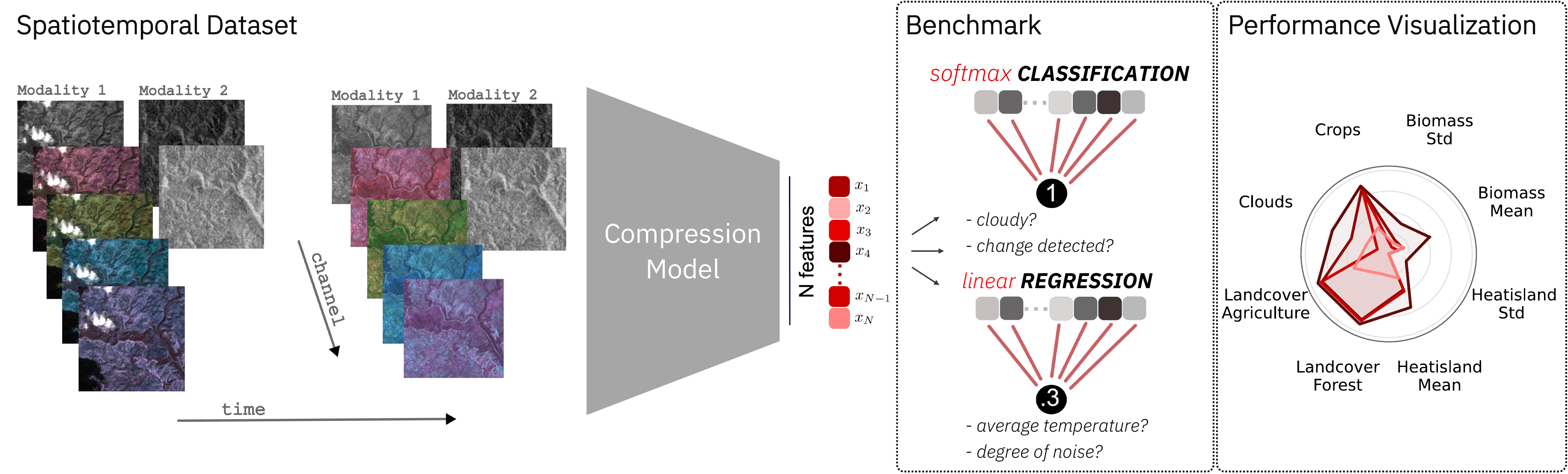}\\
    \caption{Overview of the {benchmark} framework. Multi-temporal, multi-modal, and multi-channel inputs $x$ from a spatiotemporal dataset are compressed into fixed-size embeddings $z=E(x)$ by a user-defined encoder $E$. Within the benchmark, these embeddings are linearly probed on a diverse set of regression and classification downstream tasks to assess the \textit{general-purpose} quality of $z$.}
\end{figure*}

The rapid growth of visual data, from online media to scientific observation, has made efficient compression a central challenge for storage, transmission, and large-scale analysis \citep{multimedia_bigdata,survey_multimedia,gomes2025lossy}. Traditional codecs such as JPEG2000 \citep{jpeg2000} and more recent learned autoencoders \citep{balleEndtoendOptimizedImage2016} are optimized for pixel-level distortion, largely reflecting human visual perception. However, many machine learning pipelines care less about perceptual fidelity and more about semantic fidelity, retaining the information needed to solve downstream tasks \citep{huang_machinecoding}. 
This gap is particularly critical in domains like Earth Observation (EO), where petabyte-scale datasets of multi-modal satellite imagery must support diverse analytical tasks ranging from environmental monitoring to disaster response \citep{guo_big_2017}. EO data are characterized by substantial redundancy and noise across multiple spectral bands and temporal sequences, amplifying the need for compression strategies that efficiently capture underlying, task-relevant information \citep{gomes2025lossy}. This gives rise to the  question: \emph{How much task‐relevant information can be squeezed into compact data representations?}

Recent work has shown that compressed latent representations can preserve rich semantic content, enabling pipelines to operate directly on features without reconstructing the input image \citep{torfason2018imageunderstandingdeepcompression,singh2020end}. 
Self-supervised foundation models (FMs) further demonstrate that embeddings can transfer across tasks with minimal fine-tuning. Yet, their dimensionality often rivals or exceeds the size of the original data, reintroducing storage and bandwidth bottlenecks \citep{Gomes2024NeuralEC,lu2024ai_EOFMreview}. Despite these advances, there is currently no standardized framework evaluating how effectively compressed representations retain semantic content across multiple downstream tasks. Existing evaluations remain fragmented, often restricted to pixel fidelity, single-task utility, or unconstrained high-dimensional embeddings, making it challenging to compare approaches on a common basis.

To address this, we introduce \ifthenelse{\boolean{blind}}{}{\emph{NeuCo-Bench}, } a model-agnostic benchmark for assessing the semantic quality of embeddings in EO. Our frameworks is designed to (1) evaluate compressed embeddings under strict size constraints, (2) probe semantic retention using linear models across diverse downstream tasks, (3) support multi-modal and multi-temporal data typical of data-intensive EO settings, and (4) foster community contributions, including new datasets and compressors---towards establishing open, task-centric compression standards. Our key contributions are:

\begin{itemize}[leftmargin=*,noitemsep]
\item \cref{sec:BenchmarkFramework} -- \textbf{Benchmarking Framework:}
We develop a standardized framework for evaluating compressed embeddings via downstream tasks, aligning with task-centric machine‐to‐machine workflows.
\item \cref{sec:BenchmarkTasks} -- \textbf{Benchmark Tasks:} We curate and release a suite of novel EO downstream tasks, spanning cloud analysis, agricultural monitoring, forest quantification, urban heat islands identification, and land cover analysis.
\item \cref{sec:evaluation} -- \textbf{Benchmark Evaluation:} We validate the utility of \ifthenelse{\boolean{blind}}{our benchmark}{NeuCo-Bench} through \ifthenelse{\boolean{blind}}{a}{the 2025 CVPR EarthVision} data challenge, \ifthenelse{\boolean{blind}}{}{from here on denoted the \emph{EV challenge}, }introducing a novel hidden-task evaluation scheme. We further test embedding quality under diverse compression strategies, including pre-trained neural compressors and FMs.
\end{itemize}

%% file: sects/related_work.tex
Below, we review research fields relevant to contextualize \ifthenelse{\boolean{blind}}{our benchmark framework}{NeuCo-Bench}:

\textbf{Image and neural compression.}
Image codecs such as JPEG, JPEG2000, H.264/HEVC \citep{jpeg, jpeg2000, hevc, H.264} exploit handcrafted transforms \citep{goyal_theoretical_2001, fourier, wavelets} and entropy coding to reduce redundancy. 
Learned autoencoders replace handcrafted transforms with analysis and synthesis networks jointly optimized for rate and distortion. Differentiable entropy models enable superior RD performance compared to JPEG2000 \citep{balleEndtoendOptimizedImage2016,theis2022lossy}, with subsequent extensions using hyperpriors \citep{balleVariationalImageCompression2018,autoregressive_hyperprior}, autoregressive models \citep{minnen2020channel}, and transformers \citep{qian2022entroformer}. 
Other approaches shift from human‐perceptual fidelity to task‐driven utility by jointly optimizing compressors with task networks end-to-end \citep{chamain2020endtoendoptimizedimagecompression,chamain2021endtoendoptimizedimagecompression,le2021imc,codevilla2021learnedimagecompressionmachine,wang2021endtoend,wang2023deepimagecompression,Fischer_2025}, enforcing invariance to task-relevant augmentations through self-supervised objectives \citep{dubois2022lossycompressionlosslessprediction}, or bypass reconstruction by training tasks directly on compressed latents \citep{torfason2018imageunderstandingdeepcompression,compressed_domain_inference,singh2020end}.

\textbf{Compression in EO.}
EO imagery presents unique compression challenges, with multi-spectral bands, temporal sequences, and petabyte-scale archives  \citep{guo_big_2017, wilkinson_environmental_2024}. Traditional pipelines often rely on codecs like JPEG2000 \citep{yeh_new_2005}. Recent neural approaches extend rate-distortion autoencoders to EO imagery, achieving significant rate-distortion improvements on multispectral data \citep{alves_de_oliveira_reduced-complexity_2021, kong_spectralspatial_2021, cao_spectralspatial_2022, du2024earth+, video_earth_observation}.
Importantly, most works evaluate RD, not task relevance.

\textbf{EO Foundation Models.}
Self-supervised learning has enabled vision foundation models (FMs) pretrained on vast, unlabeled satellite datasets using masked reconstruction, contrastive, or predictive tasks \citep{wang2022selfsupervisedlearningremotesensing, sun2022ringmo, wang2022advancing, Mai2022geoai, Wang23, hong2023spectralgpt, jakubik2023foundation, liu2024remoteclip}. These FMs produce versatile embeddings for EO downstream applications, such as flood segmentation, land-use mapping, and environmental monitoring. More recently, multimodal EO foundation models have begun fusing data modalities, such as SAR and optical imagery, to capture diverse geophysical characteristics and improve application performance \citep{li2022deeplearningmultimodalremote, croma, dofa,Wang_2025_ICCV_CopernicusFM, terramind, brown2025alphaearth}. However, 
the resulting dense representations often rival the original data size, creating transfer and processing bottlenecks. \citet{Gomes2024NeuralEC} address this by integrating neural compression into FM bottlenecks.
At the image level, \citet{rolf2021mosaiks} utilize fixed, random convolutional kernels to engineer features, while FM embeddings are often obtained from a CLS token or by averaging latent representations.

\textbf{Implicit neural representations.} INRs have emerged as a compelling alternative for compactly encoding Earth observation data. INR-based approaches have been explored for global location embeddings from satellite imagery \citep{klemmer2025satclip}, hyperspectral compression using neural radiance fields \citep{zhang2024compressing_,rezasoltani2024hyperspectral_}, and remote sensing image compression via coordinate-based networks \citep{li2023remote_}. More generally, INRs have shown strong potential for image compression and continuous signal representation \citep{strumpler2022implicit,dupont2021coin,sitzmann2020implicit}.

\textbf{EO benchmarks.}
Current EO domain benchmarks, such as GEO-Bench \citep{lacoste2023geobenchfoundationmodelsearth} and PANGAEA \citep{marsocci2025pangaeaglobalinclusivebenchmark}, evaluate FMs by fine-tuning backbones or training large decoders on intermediate features. These approaches typically require model access and significant computational resources, with limited consideration given to factors such as embedding size and workflow efficiency. In contrast, \ifthenelse{\boolean{blind}}{our benchmark}{NeuCo-Bench} evaluates fixed-size embeddings through task-agnostic linear probing without any need for access to model backbones. 

Our approach treats the encoder as a black box that converts inputs to a given number of features and provides a lightweight, size-aware evaluation protocol for efficient local testing. It is structured as a flexible, extendable framework, designed to accommodate future downstream tasks (e.g., \textit{Copernicus-Bench} \cite{Wang_2025_ICCV_CopernicusFM}) and evaluation methods. Moreover, \ifthenelse{\boolean{blind}}{the benchmark}{NeuCo-Bench} can be deployed as a novel challenge format that simulates real-world scenarios by requiring participants to submit EO embeddings without prior knowledge of the specific downstream tasks. This setup reflects the demand for broadly generalizable embeddings. As demonstrated in \ifthenelse{\boolean{blind}}{our data}{the EV} challenge, \ifthenelse{\boolean{blind}}{our benchmark}{NeuCo-Bench} integrates with established platforms such as EvalAI \citep{yadavEvalAIBetterEvaluation}, and is designed to support future competitions on new, unseen tasks.

%% file: sects/framework.tex
At the heart of \ifthenelse{\boolean{blind}}{the benchmark}{the NeuCo-Bench} framework resides (i) an embedding evaluation workflow and (ii) a ranking method to fairly compare performance across multiple tasks of varying difficulty.

\textbf{Evaluation workflow.}
\Cref{fig_evaluation_scheme} visualizes the pipeline for an \emph{Experiment} $p$ compressing the samples indexed by $i$ of \textit{Benchmark data} to create a set of fixed-size embeddings (\textit{Embedding set}) through an \textit{Encoder} $E^{(p)}$: These embeddings $X^{(p)}$ are provided to \ifthenelse{\boolean{blind}}{our benchmark}{NeuCo-Bench}, which performs the evaluation given corresponding \textit{Task labels} (aka \textit{downstream task}s $t=1\dots T$) to return the \textit{Benchmark results} through a \textit{Leaderboard}. For each \textit{Experiment}, the benchmark framework performs an \textit{Embedding Evaluation} given \textit{Multimodal, multitemporal datacube}s across a set of downstream tasks undisclosed to the developers of a given \textit{Compression Model} comprising an \textit{Experiment} $p$. Correspondingly, our framework aggregates scores $s_{t,k}$ per \textit{training and test split} $k$ to gather statistics for the quality score $Q_t$ per downstream task $t$. Consequently, a \textit{Scoring} algorithm applies a task difficulty-dependent ranking scheme.\ifthenelse{\boolean{blind}}{}{ A standalone Python implementation is available online\footnote{\url{https://github.com/embed2scale/NeuCo-Bench}}. 
See supplementary material for additional details.}

\textbf{Evaluating embeddings.}
For \ifthenelse{\boolean{blind}}{our benchmark}{NeuCo-Bench}, each input sample must be represented as a fixed-size embedding to compress an input data cube, e.g., our EO downstream tasks as detailed in \cref{sec:downstream_data}, or future extensions, cf.\ \cref{sec:FutureWork}. Currently, image-level linear regression and binary (softmax) classification are supported. \ifthenelse{\boolean{blind}}{Our benchmark}{NeuCo-Bench} enforces embeddings of fixed, but configurable, size but otherwise does not constrain how embeddings are generated. The embedding size can differ between tasks.
Following \cref{fig_evaluation_scheme}, the benchmark evaluates the compressed embeddings $X^{(p)}$ of an experiment $p$ as follows: For each task $t=1\ldots T$, $K$ linear classifiers (with $N$ tunable parameters $a_{1\dots N}$ plus bias term $a_0$) are trained to fit the downstream task labels $y_\text{ID}$. Each $k=1\dots K$ denotes a separate, randomly generated \texttt{split} of the downstream task $t$ into a \texttt{train}ing and \texttt{test}ing set. For each tuple $(t,k)$ the benchmark computes an accuracy measure $s_{t,k}$, utilizing $R^2$ (R-squared) for regression tasks and the $F_1$ score for (binary) classification. From the set $\left\{s_{t,k}\right\}_{k=1\dots K}$, \ifthenelse{\boolean{blind}}{our benchmark}{NeuCo-Bench} derives a signal-to-noise-like \emph{quality score} $Q_t^{(p)}$ as the mean performance on task $t$ sensitive to the variability in performance of experiment $p$:
\begin{equation}
    \label{eq_q_t}
    Q_t^{(p)} = 100\epsilon\frac{\left\langle s_{t,k}\right\rangle_k}{\text{std}_k\left(s_{t,k}\right)+\epsilon}.
\end{equation}
Here, $\left\langle\cdot\right\rangle_k$ denotes averaging and $\text{std}_k\left(\cdot\right)$ the standard deviation as calculated over the $K$ splits. The parameter $\epsilon>0$ acts as a regulator avoiding high variability in $Q_t^{(p)}$ for small $\text{std}_k\left(s_{t,k}\right)$.
The quality score \cref{eq_q_t} varies in $[0,100]$ for both, classification and regression. Thus, $Q_t^{(p)}$ allows for an interpretation of mean accuracy in percent. Compared to using the mean $R^2$ over the $K$ splits, $Q_t^{(p)}$ penalizes methods with larger variance in the $R^2$. Further details on the quality score is provided in 
the supplementary material. We note: experiments that perform worse than simply predicting the mean of labels $y_\text{ID}$ for regression tasks result in negative $s_{t,k}$ degrading the mean performance. In fact, compression models with negative $Q_t^{(p)}$ should be flagged unreliable---they seriously underperform.

\textbf{Task difficulty-dependent ranking.}
A novel scoring method is introduced by \ifthenelse{\boolean{blind}}{our benchmark}{NeuCo-Bench}. It is designed to compare the overall performance of multiple participants over multiple tasks. Based on a rank-then-aggregate approach \citep{wiesenfarth2021methods}, the benchmark dynamically weights the performance across tasks depending on their relative difficulty:  Each experiment initially receives a rank $R_t^{(p)}$ per task, with the best rank given to the experiment with highest $Q_t^{(p)}$. To break ties, all tied experiments are given the lower (better) rank. An experiment's final rank is calculated from the weighted mean rank across all tasks:
\begin{equation}
    \label{eq_weighted_rank}
    s^{(p)} = \sum_{t=1}^Tw_tR_t^{(p)}
    \quad\text{with}\quad
    w_t = \frac{\text{std}_p\left(Q_t^{(p)}\right)}{\sum_{t=1}^T\text{std}_p\left(Q_t^{(p)}\right)}
\end{equation}
where the tasks are weighted by the standard deviation of the $Q_t^{(p)}$ of all experiments on the task. The weighting scales the importance of the tasks such that (a) tasks where all participants perform similarly receive low importance, and (b) tasks where the participants differentiate between each other are weighted highly. \ifthenelse{\boolean{blind}}{Our benchmark}{NeuCo-Bench} also provides the \emph{mean $Q$} value $\langle Q_t^{(p)}\rangle_t$ as an experiment-specific measure of performance. For scenarios with few experiments where the interpretation of a ranking is limited in terms of task difficulty, $\langle Q_t^{(p)}\rangle_t$ serves as an alternative metric to compare (individual) experiments. Based on the \textit{mode of operation}, $s^{(p)}$ or $Q^{(p)}_t$ may be preferred. The former serves competitive challenge settings with novel downstream tasks of unknown levels in difficulty. The latter is favorable for long-term leaderboards where re-ranking of all methods should not depend on the update of a single method. 
Additional analysis of the ranking scheme is provided in the supplementary material.

%% file: sects/downstream_data.tex
\label{sec:downstream_data}
\ifthenelse{\boolean{blind}}{Our benchmark}{NeuCo-Bench} provides a set of pre-processed, heterogeneous EO downstream tasks designed for continuous extension in the future. The initial release provides regression labels that are easily turned into binary classification tasks through a threshold.

\begin{table*}[t]
\centering
\caption{\label{tab:downstream_overview}Summary of spatial coverage for current set of data cubes and associated number of downstream tasks.}
\label{tab:downstream_datasets}
\begin{tabularx}{0.73\textwidth}{@{} c l c c c c @{}}
\toprule
\textbf{Dataset} &
\makecell{\bf Spatial\\\bf Coverage} &
\makecell{\bf Temporal\\\bf Coverage} &
\makecell{\bf Years \\\bf of Labels} &
\makecell{\textbf{\# Samples}} &
\makecell{\textbf{\# Tasks}}\\
\midrule
{Crops} & US Corn Belt & 2022 & 2023$^\text{\ref{fn:cropyear}}$ & 3355 & 1\\ 
{Landcover} & Europe & 2018 & 2018 &  4691 & 2\\ 
{Biomass} & Global & 2019 & 2019 & 2415 & 2\\ 
{Clouds} & Global & 2018 - 2020 & 2018 - 2020 & 1140 & 1\\
{Heatisland} & Northern Hemisphere & 2022 & 2021 - 2024 &  1659 & 2\\ 
\bottomrule
\end{tabularx}
\end{table*}
\begin{figure*}[t]
    \vspace{-2ex}
    \centering
    \includegraphics[width=0.73\textwidth]{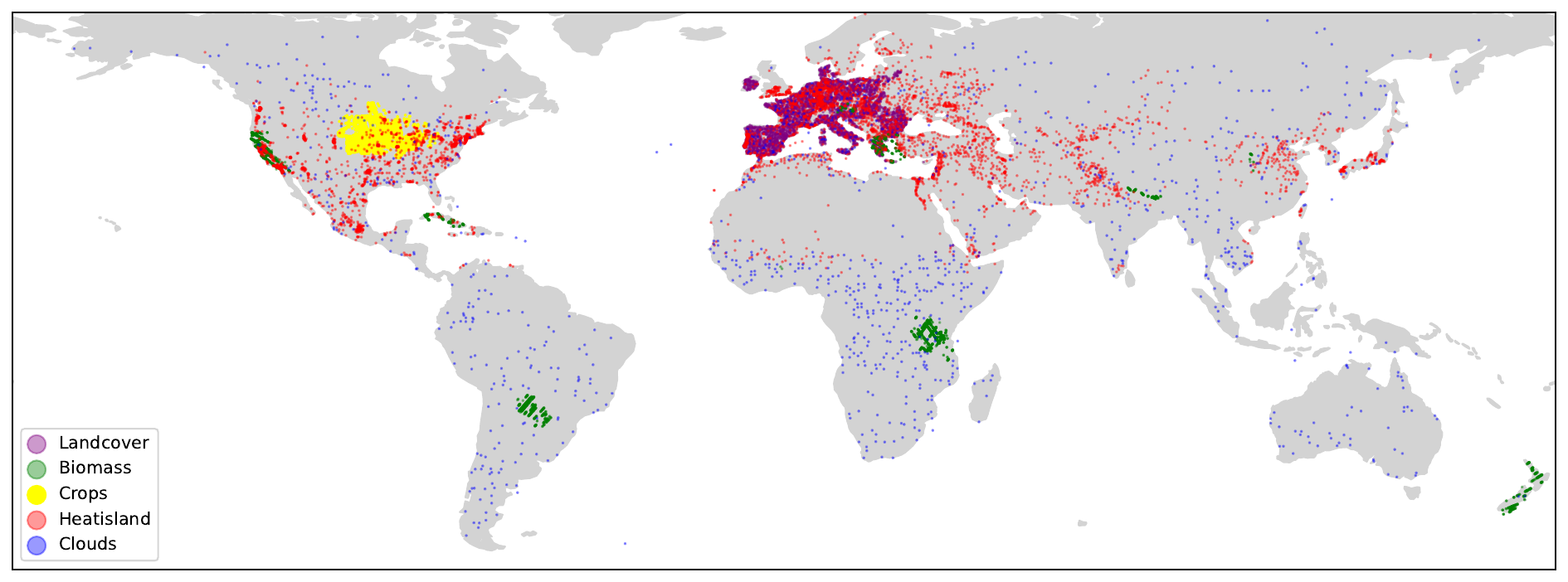}
    \caption{Spatial distribution of the downstream tasks.}
    \label{fig_neuco_bench_downstream_tasks}
    \vspace{-2ex}
\end{figure*}

The EO satellite data of the downstream tasks, summarized in \cref{tab:downstream_datasets}, follow the SSL4EO-S12 data cube structure \citep{blumenstiel2025ssl4eos12v11, Wang23}. We utilize 13 channels of Sentinel-2 Level-1C \textit{Top-of-the-Atmosphere} (S2L1C) and 12 channels of multi-spectral Sentinel-2 Level-2A \textit{surface reflectance} (S2L2A). On top we spatially align 2 channels of radar Sentinel-1 (S1) GRD product polarizations (VV and VH). For a given geolocation we retrieve four timestamps, one per season: winter (Dec--Feb), spring (Mar--May), summer (Jun--Aug), and fall (Sep--Nov). \Cref{fig_evaluation_scheme} depicts these four seasonal data cubes, with each containing 27 bands. Google Earth Engine \citep{GORELICK201718, GEE} (GEE) was utilized to download all relevant satellite data. All labels, except for the Clouds use case, have been retrieved from GEE, too. 
The processed data was stored as cloud-ready ZIP-Store of the Zarr file format.

The downstream tasks contain between 1100 and 4691 samples (locations/labels), which are distributed globally (\cref{fig_neuco_bench_downstream_tasks}). The associated satellite data cubes are pre-processed and filtered to ensure UTM-projected patches with a size of 264 × 264 pixels without spatial overlap.

The \textbf{crops} task covers cropland in the US Corn Belt and is provided by the US Department of Agriculture (USDA) \citep{Boryan_CDL}. Soybean and corn were selected as primary focus classes for this downstream task, with the fraction of corn and soybean within each patch serving as the label. The Crop Data Layer is published annually with a spatial resolution of 30 meters. The labels are available as post-processed data\footnote{\label{fn:cropyear} e.g., the label year 2023 corresponds to crops cultivated during the 2022 growing season}.

The \textbf{landcover} tasks leverages aggregated land use data from the European Environment Agency (EEA), which includes various land cover classes such as forests, urban areas, water areas, and agricultural land \citep{CLC2018}. These labels represent the dominant land cover within each patch at a spatial resolution of 100 meters within Europe. Based on this data, two downstream tasks are provided for forests and agricultural land 2018 within the challenge.

The \textbf{biomass} tasks uses above-ground biomass estimates derived from LIDAR measurements from the Global Ecosystem Dynamics Investigation (GEDI) instrument, and are sampled based on \citep{Sialelli2025}. GEDI provides structural information on vegetation height and density, allowing robust models to estimate above-ground biomass in megagrams per hectare (Mg/ha) \citep{GEDI}. For the labels, GEDI Level 4A biomass estimates are spatially aggregated to the satellite patches with 264x264 pixels, providing a mean biomass value and its standard deviation as regression targets.

The \textbf{clouds} provides cloud cover fractions based on CloudSen12+ \citep{Ayabar2024CloudSen12} as labels and pre-processed Sentinel-1 and Sentinel-2 data cubes as corresponding observations. Although the SAR data is not affected by clouds, Sentinel-1 is included alongside Sentinel-2 to ensure a consistent data structure for all downstream tasks.

The \textbf{heatisland} task uses the spatio-temporal mean and standard deviations from Landsat-8 \citep{landsat8} Land Surface Temperature (LST) of urban areas as labels. This is particularly relevant in the context of heat events and future urban planning.

Further information on the data and their downstream tasks can be found in the supplementary material.

%% file: sects/evaluations_cvpr.tex
\subsection{Data Challenge Validation}
\label{sec:exp_setup}
To validate \ifthenelse{\boolean{blind}}{our benchmark}{NeuCo-Bench} under realistic conditions, we utilized it in \ifthenelse{\boolean{blind}}{a}{the 2025 CVPR EarthVision} data challenge. Participants were tasked with compressing multi‐modal, multi‐temporal EO imagery, cf.\ \cref{sec:downstream_data}, into 1,024-dimensional embeddings, corresponding to a compression ratio of approximately 7,000.
Crucially, participants did not know which or the number of downstream tasks their embeddings would be evaluated on; this hidden‐task design discourages overfitting and encourages the development of general‐purpose EO representations.
Participants were ranked, in accordance with \cref{sec:BenchmarkFramework}, across two sets of downstream tasks. One modification was made to the dataset compared to the dataset described in \cref{sec:downstream_data}; the clouds task targets were mainly replaced by zeros, causing heavy skewness in the labels and basically random connection between the imagery and labels.

\textbf{Phases.} In a three‐week development phase, teams developed embedding methods using \ifthenelse{\boolean{blind}}{a}{the} publicly available \ifthenelse{\boolean{blind}}{}{SSL4EO-S12 v1.1 }dataset. A partial release of 5 tasks, each with a subset of its samples but no information of the task type, allowed participants to receive initial feedback for development.  Submissions returned only the mean $Q$ value to prevent leakage of task‐specific performance and information.

In the subsequent three‐day evaluation phase, an extended set of 9 tasks, and new data on the tasks also used in the development phase, was released. The teams had three days and up to three submissions to encode and submit embeddings; these runs defined the final leader board standings. By the end, two winning teams were chosen; The first based on the dynamic ranking scheme, and the other as the team with highest mean $Q$ score.

\textbf{Platform and infrastructure.} The \ifthenelse{\boolean{blind}}{benchmark}{NeuCo-Bench} framework was modified to utilize Eval.AI \citep{yadavEvalAIBetterEvaluation} to collect submissions. The benchmark ran on a separate 8-vCPU server, retrieved new submissions via API, executed the evaluation, and pushed results to a custom leader board hosted on GitHub, displaying the dynamic ranking described in \cref{sec:BenchmarkFramework}, as well as back to Eval.AI. Additional details are available in the supplementary material.

\subsection{Data Challenge Results}
\label{sec:eval_cvpr}
\begin{figure}[!ht]
    \centering
    \vspace{-3mm}
    \includegraphics[width=0.9\linewidth]{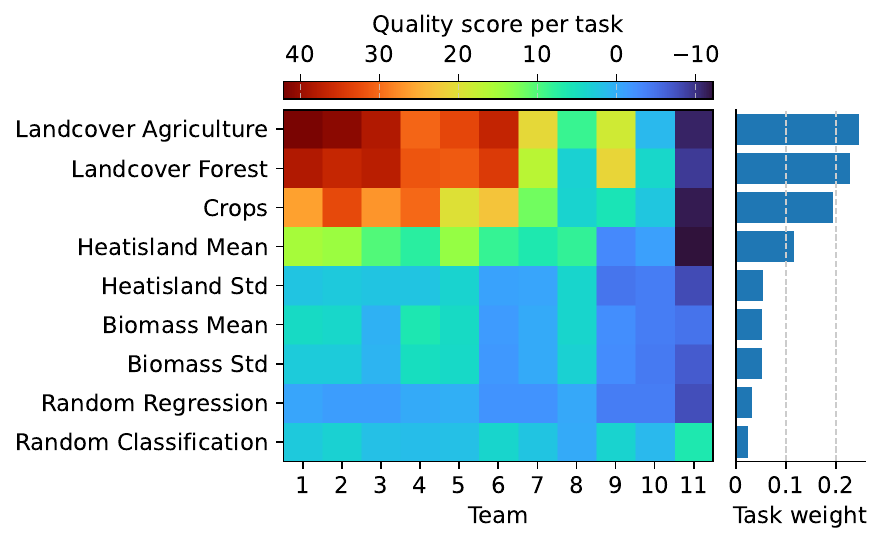}
    \caption{Quality score $Q_t^{(p)}$ of the participants of the data challenge evaluation phase with corresponding task weight used for ranking. Teams are ordered by final leader board rank (Team~1 is the winner). Team 10 is a simple averaging baseline (\cref{sec:eval_general}) and Team~11 is a normally distributed random embeddings baseline. We observe that landcover tasks are the most discriminative, the random control task is the most ``difficult''(by design), and estimating biomass 
    is the most challenging real-world downstream task.}
    \label{fig:data_challenge_performance}
    \vspace{-4mm}
\end{figure}

\textbf{Participation and ranking.}
Twenty‐three teams submitted to the development phase; sixteen went on to the final evaluation, nine of which shared their submissions publicly. The quality scores \(Q_t^{(p)}\), shown in \cref{fig:data_challenge_performance} for the evaluation phase, varied widely from 0--5 on some tasks to 5--40 on others. The evaluation method, cf.\ \cref{eq_weighted_rank}
, efficiently scaled task importance ensuring that the tasks impacted the leader board relative to their differentiating effect across the teams. Notably, the weighting reduced the impact of the tasks with random labels. Further, the dynamic ranking caused a swap between the original first and second place methods due to a third team, highlighting the impact of adaptive weighting. Detailed specifications of the challenge evaluations is given in the supplementary material.

\textbf{Top methods.}
The team that achieved the best overall rank and the team with the highest average \(Q\)-Score both built their embeddings by ensembling multiple FM representations, although with different approaches; One pre-training backbones, and the other training a bottleneck based on frozen FM backbones. The fourth-place team took a different path, forgoing any pre-training and instead generating embeddings using the MOSAIKS method \citep{rolf2021mosaiks}.

\textbf{Key takeaways.} \ifthenelse{\boolean{blind}}{Running the data challenge}{The EV challenge} demonstrated that \ifthenelse{\boolean{blind}}{our benchmark}{NeuCo-Bench} efficiently evaluates and ranks the performance of compact embeddings over multiple downstream tasks. The scoring method produced a more balanced and discriminative ranking compared to uniformly weighting the tasks, particularly noticeable in the down-weighting of the two random tasks. Hiding the tasks efficiently prevented overfitting, ensuring fairness between the participants. Both winning solutions were based on FMs, indicating these can indeed provide semantically rich, general embeddings. However, also non-FM based solutions scored high.

%% file: sects/evaluations_general.tex
\subsection{General Evaluations}
\label{sec:eval_general}
We assess \ifthenelse{\boolean{blind}}{our}{the NeuCo-Bench} framework through a series of experiments, including embeddings from self-supervised FMs and representations from learned neural compressors. To meet the compression requirements, we apply spatial and temporal aggregation techniques to obtain compact embeddings.

Our analyses span three perspectives: (i) we follow the challenge setup and constrain all methods to 1,024-dimensional embeddings; (ii) we relax the embedding dimension constraint to study how embedding dimensionality affects downstream performance, motivating the challenge default; and (iii) we revisit the assumption of linear probes replacing them by more complex decoder models.

\textbf{Embedding Aggregation Methods.}
Each input consists of four seasonal snapshots from Sentinel-1 (radar) and Sentinel-2 (optical), as described in \cref{sec:BenchmarkTasks}.
We benchmark image-based encoders by applying temporal averaging across the four timesteps, either \emph{before} encoding (pre-encoding aggregation) or \emph{after} encoding (post-encoding aggregation). 
Convolutional encoder outputs are aggregated through spatial averaging in embedding space, with additional pairwise channel means applied when further dimensionality reduction is needed. For ViT encoders, we either average the spatial patch tokens to obtain a single embedding or use the CLS-token.
Aggregated embeddings smaller than the target size (e.g., 1024) are zero-padded. To study the role of input modalities, we evaluate unimodal encoders that use only S2L1C bands and multimodal encoders that process the full SSL4EO-S12 data cube.

\begin{table*}[t]
  \centering
  \small
  \caption{Full per‐task $R^2$ scores for tested embedding methods including average performance (Avg.) across all tasks. For ViT's, experiments utilizing the CLS token for prediction is indicated by (CLS), otherwise the average patch token is used. Methods are sorted in ascending order by Avg. For each task, the best-performing method is highlighted in \textbf{bold}, and the second‐best method is \underline{underlined}. Abbreviations: Landcover (LC), Heatisland (HI), Agriculture (Agri), Average (Avg.) across all downstream tasks \vspace{-2ex}}
  \label{tab:full_results}
  \begin{tabular}{lrrrrrrrr|r}
    \toprule
    \textbf{Method}
      & \makecell{\textbf{Biomass}\\\textbf{Mean}}
      & \makecell{\textbf{Biomass}\\\textbf{Std}}
      & \textbf{Crops}
      & \textbf{Clouds}
      & \makecell{\textbf{LC}\\\textbf{Agri}}
      & \makecell{\textbf{LC}\\\textbf{Forest}}
      & \makecell{\textbf{HI}\\\textbf{Mean}}
      & \makecell{\textbf{HI}\\\textbf{Std}}
      & \textbf{Avg.} \\
    \midrule
    Averaging Baseline
    & -0.552 & -0.426 & -0.061 & -2.293 & 0.126 & 0.216 & -0.962 & -1.157
    & -0.514 \\
    \midrule
    Factorized Prior ($\lambda$ = 0.5)
    & -0.036 & -0.071 & 0.325 & 0.200 & 0.478 & 0.343 & 0.033 & -0.628
    & 0.080 \\
    
    Factorized Prior ($\lambda$ = 0.1)
    & 0.129 & 0.078 & 0.357 & 0.266 & 0.464 & 0.333 & 0.219 & -0.270
    & 0.197 \\
    
    Factorized Prior ($\lambda$ = 0.025)
    & 0.195 & 0.140 & 0.338 & 0.288 & 0.449 & 0.256 & 0.315 & -0.113
    & 0.233 \\
    \midrule

    S12-DINO ResNet50~\citep{Wang23}
    & 0.117 & 0.088 & 0.826 & 0.147 & \underline{0.879} & \underline{0.865} & 0.483 & -0.221
    & 0.286 \\
    
    S12-MoCo ResNet50~\citep{Wang23}
    & -0.048 & -0.030 & 0.780 & 0.216 & 0.864 & 0.844 & 0.345 & -0.334
    & 0.286 \\
    
    Satlas MS ResNet50~\citep{satlas},
    & 0.310 & 0.248 & 0.490 & 0.340 & 0.658 & 0.641 & 0.485 & 0.092
    & 0.408 \\
    
    Clay-V1 ViT-B (CLS)~\citep{dofa}
      & 0.199
      & 0.169
      & 0.748
      & 0.482
      & 0.811
      & 0.803
      & 0.430
      & -0.013
      & 0.454  \\
  
    S12-DINO ViT-S~\citep{Wang23}
    & 0.184 & 0.183 & \underline{0.842} & 0.484 & 0.863 & 0.851 & 0.461 & -0.109
    & 0.470 \\
    
    S12-MoCo ViT-S~\citep{Wang23}
    & 0.338 & 0.259 & 0.751 & 0.409 & 0.825 & 0.814 & 0.506 & 0.133
    & 0.504 \\
    
    DOFA-ViT-L~\citep{dofa}
    & 0.373 & 0.269 & 0.587 & 0.518 & 0.777 & 0.773 & 0.560 & 0.207
    & 0.508 \\
    
    S12-MAE ViT-S~\citep{Wang23}
    & 0.380 & 0.280 & 0.627 & 0.659 & 0.792 & 0.797 & 0.585 & 0.181
    & 0.537 \\
    
    Prithvi-EO-V2-300M (CLS)~\citep{prithvi}
      & 0.426
      & 0.329
      & 0.655
      & 0.531
      & 0.764
      & 0.782
      & 0.603
      & 0.216
      & 0.538 \\

    Prithvi-EO-V2-300M~\citep{prithvi}
    & 0.424 & 0.324 & 0.671 & 0.588 & 0.781 & 0.792 & \underline{0.608} & 0.203
    & 0.549 \\

    Clay-V1 ViT-B~\citep{dofa}
    & \underline{0.466} & \underline{0.332} & 0.759 & \underline{0.660} & 0.825 & 0.825 & 0.606 & \textbf{0.229}
    & \underline{0.588} \\

    TerraMind-V1-B~\citep{terramind}
    & \textbf{0.528} & \textbf{0.390} & \textbf{0.879} & \textbf{0.731} & \textbf{0.918} & \textbf{0.908} & \textbf{0.691} & \underline{0.226}
    & \textbf{0.659}  \\ 
  
    \bottomrule
  \end{tabular}
\end{table*}

\textbf{Encoder Baselines.} A simple \textit{averaging baseline}, which applies spatial pooling, channel‐wise averaging, and flattening serves as a minimal reference point, see the supplementary for details.
Neural rate–distortion compressors are implemented via Factorized Prior autoencoders~\citep{balleEndtoendOptimizedImage2016} pretrained on \ifthenelse{\boolean{blind}}{}{SSL4EO }S2L1C data. We extract latent bottleneck features before entropy coding (cf.\ \citep{torfason2018imageunderstandingdeepcompression}) and aggregate as described above.
We benchmark a range of publicly available EO FMs, including unimodal ResNet \citep{resnet} and ViT \citep{vit} backbones, 
and multimodal models like TerraMind~\citep{terramind} and DOFA~\citep{dofa}.

\textbf{Results and discussion.} 
\Cref{tab:full_results} presents the linear‐probing performance in terms of $R^2$ for all downstream tasks.
Embeddings from neural rate–distortion compressors outperform the simple averaging baseline but generally remain below $R^2=0.5$. This reflects the characteristics of our setup which combines high compression rate $\sim$7,000 with linear probing. FM embeddings show a task‐dependent trend: FMs in general, with contrastive (DINO, MoCo) and multimodal models (TerraMind) in particular, achieve high $R^2$ on semantic tasks where multi-pixel context is relevant (e.g., land‐cover proportion). However, certain FMs struggle on geophysical predictions of quantities resolved at the sub-pixel level (e.g., biomass estimation).

\textbf{Temporal aggregation.}
Across all methods, post‐encoding aggregation consistently outperforms pre‐encoding aggregation (see the supplementary). Accordingly, the results in \Cref{tab:full_results} all use temporal post‐encoding aggregation. The gains are modest for static‐feature tasks (e.g., land‐cover), yet substantial for temporally sensitive tasks (e.g., cloud‐fraction estimation), underscoring the importance of preserving per‐snapshot details.

\begin{figure}[ht]
  \centering
  \includegraphics[width=0.85\linewidth]{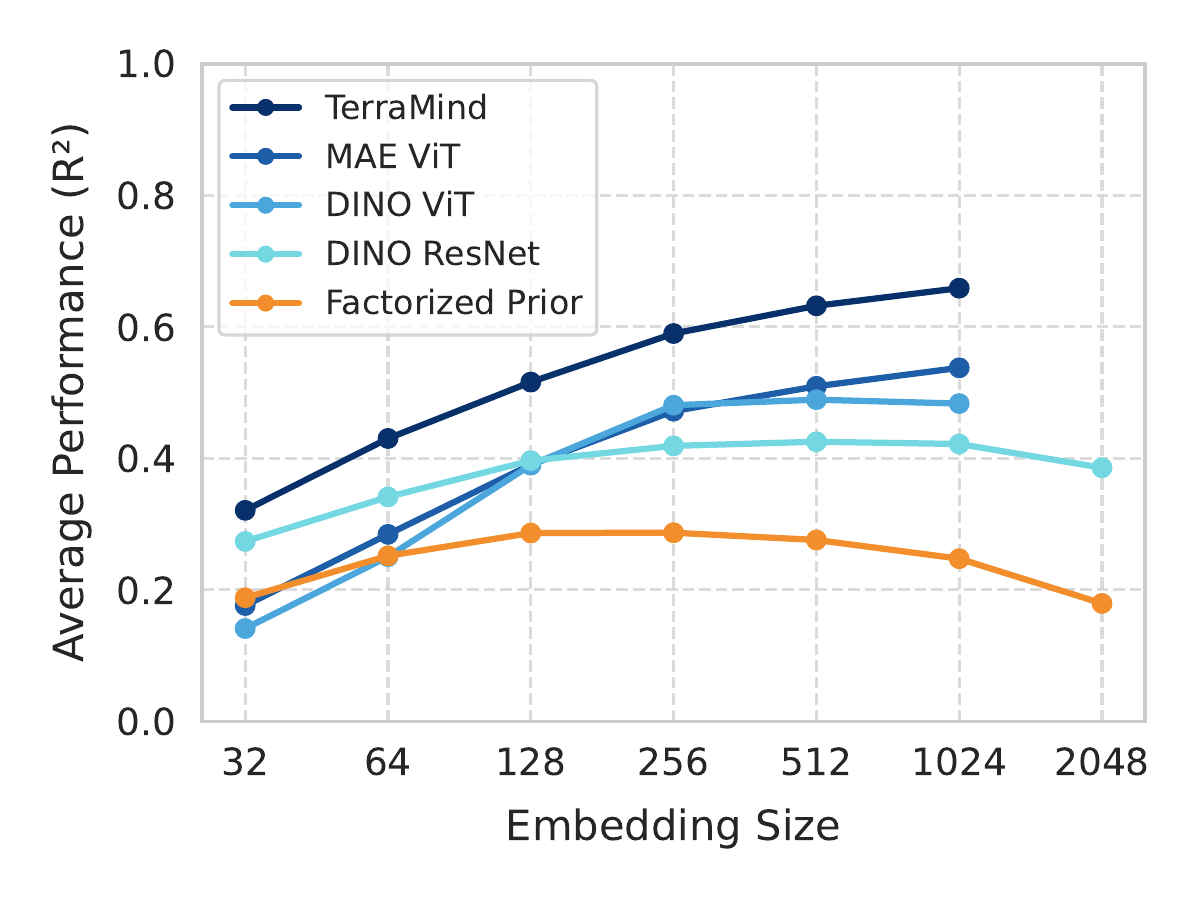}
  \vspace{-4mm}
  \caption{Average task $R^2$ vs. embedding size $\log N$. Largest size is the full channel count (CNN) or patch-token dim. (ViT).}
  \label{subfig:embedding_size}
  \vspace{-4mm}
\end{figure}

\textbf{Embedding size.}  
In \cref{subfig:embedding_size} we study numerically how embedding size impacts performance on a subset of models, with per-task results in the supplementary material. Given our setup, we observe:
\begin{itemize}
    \item CNN backbones. Performance peaks for $128\lesssim N\lesssim1024$ and drops outside this range. Larger embeddings add computational cost without significant performance gains.  
    \item ViT backbones. We find the best performance at $N=1024$, the upper limit allowed by the embedding dimension. A lower $N$ consistently reduce accuracy---except for certain regression tasks such as for \textit{Biomass}.  
    \item Trade-offs. While larger embeddings increase the number $N=\vert A\vert$ of (linear) probe parameters $A$ (cf.\ \cref{fig_evaluation_scheme}),
    smaller $N$ often fail to retain task-relevant semantics.  
\end{itemize}
Our experiments support an embedding size of $N=1024$ as a practical default across tasks, cf.\ \cref{fig_neuco_bench_downstream_tasks}. The \ifthenelse{\boolean{blind}}{benchmark}{NeuCo-Bench} framework is flexible to explore size–utility trade-offs, analogous to rate–distortion analysis in neural compression.

\textbf{Linear probing assumption.}  We evaluate embedding quality using linear probing, a standard practice in representation learning \citep{Ziping2021}, to assess embeddings without fine-tuning encoder backbones. While non-linear probing (e.g., small MLP heads) can capture richer structures, they can compensate for poor embedding quality~\cite{plachouras2025unifiedrepresentation}. 

\begin{figure}[ht]
  \centering
  \includegraphics[width=0.85\linewidth]{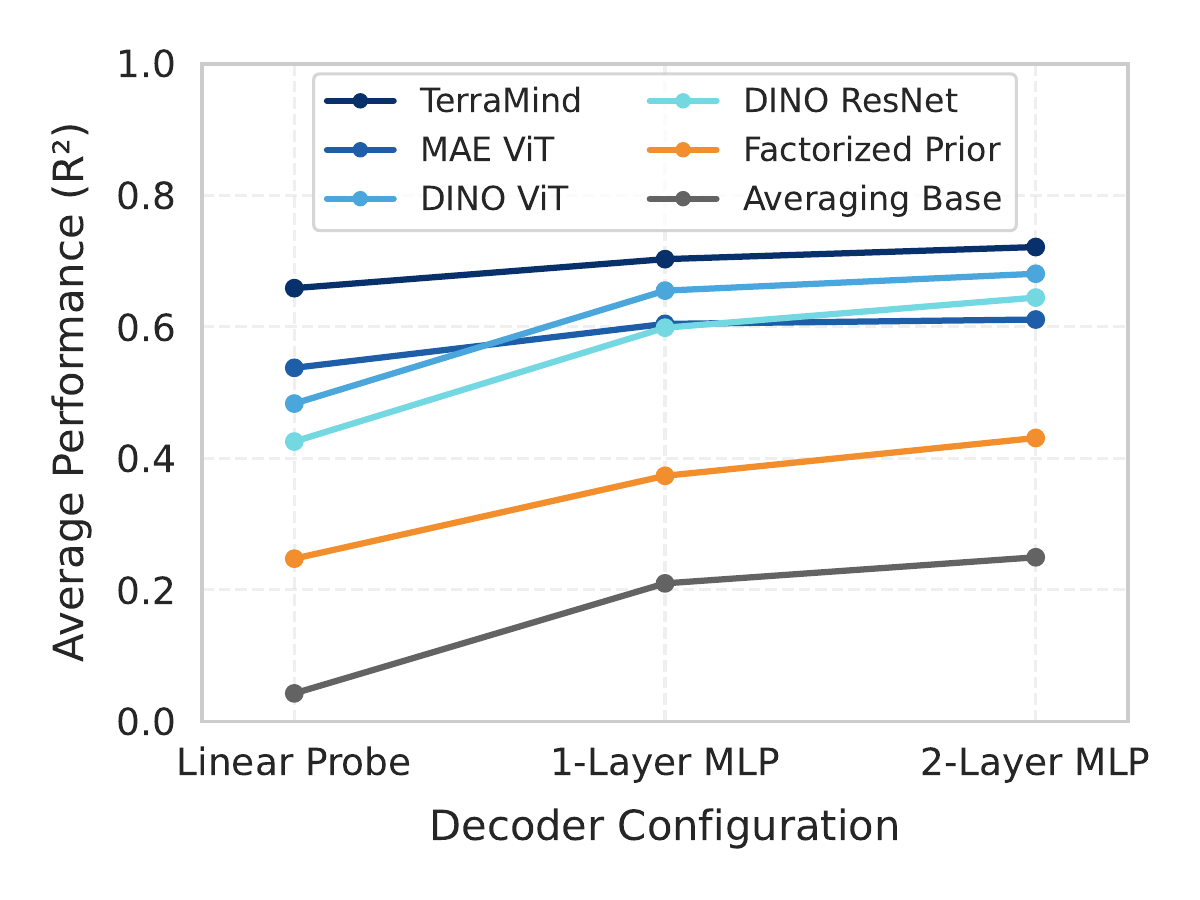}
  \vspace{-4mm}
  \caption{Average $R^2$ comparing linear probing against one-layer and two-layer MLP probes on 1024-dimensional embeddings.}
  \label{subfig:decoder_size}
  \vspace{-2mm}
\end{figure}

Our experiments in \cref{subfig:decoder_size} and 
with additional details in the supplementary material show that replacing linear probes by small, non-linear decoders yields only marginal gains for top-performing embeddings, while providing larger improvements for weaker ones, at substantially higher computational cost. Linear probing remains an efficient and reliable measure of how much semantically relevant information is directly accessible from an embedding space. Its efficiency enabled \ifthenelse{\boolean{blind}}{our benchmark}{NeuCo-Bench at the EV challenge} to run over 400 submissions within minutes on commodity hardware.

%% file: sects/limitations_futurework.tex
\section{Future Work}\label{sec:FutureWork}
\textbf{Reproducibility.} 
Data contributions to \ifthenelse{\boolean{blind}}{our benchmark}{NeuCo-Bench} demands a permissive CC-BY 4.0 license. Our data challenge required the winners to release their solution under Apache 2.0 license\footnote{\relsize{-1.5}\url{https://creativecommons.org/licenses/by/4.0} and \url{ https://www.apache.org/licenses/LICENSE-2.0.html}}. The same holds for any future extension of \ifthenelse{\boolean{blind}}{the benchmark}{NeuCo-Bench}. 
Details how to extend the framework, reference code, hardware specifications, software environments and hyperparameters used in the data challenge are provided in the supplementary material.

\textbf{Fixed-size compression.}  
Our current evaluation emphasizes fixed-size embeddings, as fixed-size vectors enable fast retrieval, comparison, and inference, critical for machine-oriented downstream tasks. Nevertheless, the framework can be naturally extended to incorporate entropy coding, where embeddings are further losslessly compressed for transmission before decoding and use. In this setting, the proposed performance scores directly evaluate utility as a function of the final entropy-coded bitrate, thereby bridging task performance with classical rate–distortion analysis.

\textbf{Choice of tasks.} 
The discriminative power of results depends on downstream tasks. We curated diverse, image-level tasks focused on global semantic content, together with a dynamic ranking scheme. Future extensions will include spatially structured tasks such as pixel-level segmentation or time-sensitive predictions, which require less aggressive compression ratios than the current value of $\sim$7,000. Probing strategies may also evolve as tasks grow in complexity.

\textbf{Downstream data.}  
Although the current benchmark is rooted in EO, its design is domain-agnostic. Extensions can cover multi-modal, spatio-temporal data across domains such as weather forecasting, medical imaging, or autonomous driving. Our experiments leveraged SSL4EO-S12 as an initial sweet spot (multi-modal, multi-temporal, multi-spectral). The same concept readily transfers along the axes of \textit{domain}, \textit{modality}, \textit{time}, and \textit{channels}.

\textbf{Building a \ifthenelse{\boolean{blind}}{}{NeuCo-Bench }community.}
With \ifthenelse{\boolean{blind}}{our benchmark}{NeuCo-Bench}, we provide a seed to grow an ecosystem centered around benchmarking compact embeddings on a set of standardized, community-contributed downstream tasks.
To avoid contributing to an ever-growing number of benchmark datasets, we intend to harmonize with existing ones, such as GEO-Bench and PANGAEA. 
Future extensions of the benchmark framework include pixel-wise and temporal downstream tasks.

\ifthenelse{\boolean{blind}}{}{%
\textbf{Societal impact \& information privacy.}
\ifthenelse{\boolean{blind}}{Our benchmark}{NeuCo-Bench} fosters open science for neural compression. 
SSL4EO-S12 derives from the Copernicus Programme under a free and open data policy. For label sources, please check details in 
the supplementary material.
The provided dataset is void of detailed time and geolocation information, sparsely spread in time with a single timestamp per quarter, and limited to 10m pixel resolution, limiting any surveillance applications. For benchmarks with strong compression ratio, \ifthenelse{\boolean{blind}}{our benchmark}{NeuCo-Bench} may act as a privacy conserving approach where pixel-level reconstruction becomes impossible. \ifthenelse{\boolean{blind}}{Our benchmark}{NeuCo-Bench} does not touch on any concerns related to human rights. We include the Linux Foundations \textit{Code of Conduct} for future contributions to the code released with this paper.
}

%% file: supplement-contents.tex
\section{Technical Details of the Data \mbox{Challenge}}

Before we delve into generic considerations regarding \ifthenelse{\boolean{blind}}{our benchmark}{NeuCo-Bench} in \cref{sec:neucobench_framework}, we introduce its origin spared by the innocent question:
\begin{quote}
    \it If Geospatial Foundation Models claim to generate informative, generic feature vectors for a broad range of use cases, why can't we put that claim to the test in a data challenge\ifthenelse{\boolean{blind}}{}{ at this year's CVPR}? Catch: We will not disclose the downstream tasks, but simply ask to embed \slash compress Earth observation data.
\end{quote}

\subsection{Challenge Evaluation Method and Configuration}\label{sec:sup_challenge_eval}

\textbf{An individual, \textit{local} score of embedding quality.} As detailed in the main article, the central evaluation metric serving as quality score to answer the question above reads like
\begin{align}
    \label{eq:quality_score}
    Q_t^{(p)}
    = 100\epsilon\frac{\left\langle s_{t,k}\right\rangle_k}{\text{std}_k\left(s_{t,k}\right)+\epsilon}
    &\equiv Q
    = 100\epsilon\frac{\bar s}{\Delta s+\epsilon}\\
    \nonumber
    \text{with}\quad\epsilon&=0.02.
\end{align}
For fixed $\epsilon>0$, the maximum value of $Q$ reaches $100\epsilon\epsilon^{-1}\bar s=100\bar s$ when the statistical fluctuations vanish, $\Delta s\to0$. Given $\Delta s\ge0$ and $\bar s\in[0,1]$\footnote{For an $R$-squared score (regression task), $s<0$ penalizing good, positive values $s\in[0,1]$. In fact, negative $s$--values indicate that the downstream task prediction is worse than a model simply predicting the value of the mean label.} derived from a measure such as the F1-score or R-squared, the range of the quality score can be interpreted as a \textit{percentage of quality}.

The numerical value of the regulator $\epsilon$ determines the scale at which $Q$ becomes insensitive to statistical fluctuations $\Delta s$: As long as $\Delta s\gg\epsilon$, in zero-order approximation, we have $q=Q/100\approx\bar s/\Delta s$ a measure of signal-over-noise for the quantity $s$.  At the other end of the spectrum when $\epsilon\gg\Delta s$ dominates the noise, we conclude $q\approx(1-\Delta s/\epsilon)\bar s<\bar s$ in first order of $\Delta s/\epsilon$. However, when $\Delta s\approx\epsilon$, $q\approx\bar s/2$ is relatively insensitive to the noise $\Delta s$. In particular, when the score $s$ varies about $\Delta s\approx0.02=2$\%$\approx\epsilon$ across the set of validations indexed by $k$, then $Q\approx50\bar s$, i.e. for almost perfect $s\approx1$ values across the board, we obtain a $Q$ close to $50$\%. Only, when $\Delta s$ significantly drops below the fixed $\epsilon=2$\%, $Q$-scores close to 100\% are possible (given close to perfect $s$-scores). 

In order to gather sufficient statistics to fairly compare the challenge participants, the number of linear classifiers trained on separately-sampled training and test sets was varied from $k=1,2,\dots,40$ during the development phase and $k=1,2,\dots,200$ during the final evaluation phase. While the seed for the random number generator used for the training and test set splits is kept constant for \ifthenelse{\boolean{blind}}{our benchmark}{NeuCo-Bench} in \cref{sec:neucobench_framework}, for the \ifthenelse{\boolean{blind}}{}{CVPR EARTHVISION }data challenge it was initialized at random. Our choice was motivated by the effort to minimize information leakage about the hidden downstream tasks to the data challenge participants. During the development phase, submissions could test the constant set of predefined downstream tasks over a three-week period and submit 10 times a day.

\textbf{\textit{Global} ranking relative to other challenge participants.} On top of a single participants $p$'s (\textit{local}) performance score $Q_t^{(p)}$, we added a \textit{global} ranking scheme as follows: Both local and global rankings assign rank $R_t^{(p)}$=1 to the highest performing participant and ascending rank $R_t^{(p)}$-values for decreasing performance. Ties are broken such that all tied participants get the lower (best) rank. The algorithmic design of our approach is best illustrated in a Python code implementation like:
\begin{lstlisting}[style=PythonStyle]
q = {
    'team1': 13.223,
    ...,
    'teamP': -3.55677
}

def rank(q:dict, descending:bool = True) -> dict:
    sign = 1
    if descending:
        sign = -1
    return {
        p: 1 + len(
             [ s_sub for s_sub in q.values() 
               if sign*s_sub < sign*s ]
           ) 
        for p, s in q.items()
    }
    
ranked_q = rank(q)
\end{lstlisting}
where the Python dictionary \texttt{q} serves as input to \texttt{rank()} to generate $R_t^{(p)}$=\texttt{ranked\_q} and the boolean parameter \texttt{descending} triggers whether the highest or lowest value is deemed best. Utilizing \texttt{rank()}, the local ranking $R_t^{(p)}$ orders participant $p$ on task $t$ with the highest (best) score $Q_t^{(p)}$, \texttt{descending=True}. The second, global ranking across tasks assigns rank $R^{(p)}=$1 to the participant with the lowest (best) weighted average rank score

\begin{align}
    \label{eq:global_rank_score}
    s^{(p)} &= \sum_{t=1}^Tw_tR_t^{(p)}\\
    \nonumber
    \text{where}\quad
    w_t &= \frac{\text{std}_p Q_t^{(p)}}{\sum_{t=1}^T\text{std}_p Q_t^{(p)}}
\end{align}
by setting \texttt{descending=False}. In contrast to $\text{std}_k$ over cross-validation folds $k$ in \cref{eq:quality_score}, here, $\text{std}_p$ runs over the number of participants $p$ of a fixed task $t$, i.e., the weight $w_t$ computes the variation of our evaluation metric $Q^{(p)}_t$ for a given task $t$ across all data challenge participants $p$. Thus, $w_t$ serves as a measure of \textit{task competitiveness} to characterize and automatically distinguish tasks $t$.

Our design rationale of the \texttt{weighted\_score} for the \ifthenelse{\boolean{blind}}{}{2025 CVPR EARTHVISION }data challenge was as follows:
\begin{itemize}
    \item reward participants scoring well for a given downstream task
    \item discount the quality score $Q$ depending on the \textit{task competitiveness} of a downstream task, i.e., measure relative performance among challenge participants for a given downstream task.
\end{itemize}
The $\text{std}$-based weighting achieves this by discounting downstream tasks where all teams perform similarly, in analogy to:
\begin{quote}\it
A football match is a draw regardless if the end result is 1-1 or 8-8 --- although the number of goals can have a marginal effect in a tournament.
\end{quote}
We assign more importance to downstream tasks where participants score high AND when they distinguish themselves from the rest. More formally speaking: For a weight $w_t = \delta_t/\sum_\tau\delta_\tau$ with $\delta_t = \text{std}_p Q^{(p)}_t$ and the commonly accepted definition of variance
\begin{equation}
(\text{std}_p A_p)^2 = \langle A_p^2\rangle_p-\bar A^2 = \langle(A_p-\bar A)^2\rangle_p
\end{equation}
where
\begin{equation}
\langle f(X_p)\rangle_p = \frac{1}{P+1}\sum_{p=0}^P f(X_p)
\quad\text{and}\quad
\bar A = \langle A_p\rangle_p
\end{equation}
such that
\begin{equation}
A_p=\bar A\quad\Leftrightarrow\quad\text{std}_pA_p=0~, 
\end{equation}
the case $\delta_t\to0$ for all $t$ may generate a numerical instability. However, our two distinct competition baselines
\begin{itemize}
    \item $p=1$: simple data aggregation of \ifthenelse{\boolean{blind}}{}{SSL4EO-S12 }data cubes termed \textit{Baseline mean embeddings} in the \ifthenelse{\boolean{blind}}{}{2025 CVPR EARTHVISION }data challenge with leaderboard mean Q-score $\langle Q_t^{(1)}\rangle_t=$\texttt{-0.786}
    \item $p=0$: random embeddings termed \textit{Baseline random embeddings} in the \ifthenelse{\boolean{blind}}{}{2025 CVPR EARTHVISION }data challenge with leaderboard mean Q-score $\langle Q_t^{(0)}\rangle_t=$\texttt{-7.092}
\end{itemize}
prevent $\delta_t=0$ in practice as verified by running the \ifthenelse{\boolean{blind}}{}{CVPR EARHTVISION }data challenge over a month with more than 400 submissions from over 20 teams.
 
From a theoretical perspective, one may want to stabilize $w_t$ by adding a \textit{ghost task} $t=0$ with variance $0<\delta_0=\epsilon\ll1$ such that
\begin{align}
\delta_0&=\sqrt{\left\langle Q_0^{(p)2}\right\rangle_p}>0\\
\nonumber
\text{setting}\quad
\bar Q_0&=0
\quad\text{and defining}\quad
R_0^{(p)}=0~.
\end{align}
Abbreviating $\sum=\sum_t\delta_t$ we distinguish the cases
\begin{itemize}
    \item $\sum\gg\epsilon$: where $w_0=\epsilon/\sum\ll1$ and $w_t=\delta_t/\sum$ leaving $s^{(p)}$ of \cref{eq:global_rank_score} intact to 0th order in $\epsilon$
    \item $\sum\approx\epsilon$: where $w_0\approx\frac{1}{2}$ and $w_t\approx\frac{1}{2}\delta_t/\sum$ such that with $R_0^{(p)}=0$ the score $s^{(p)}$ in \cref{eq:global_rank_score} receives a discount factor $\tfrac{1}{2}$ which further increases for $\sum\to0$ where $w_0\to1$
\end{itemize} 

For the \ifthenelse{\boolean{blind}}{}{CVPR EARTHVISION }data challenge we ran \ifthenelse{\boolean{blind}}{our benchmark}{NeuCo-Bench} with task weighting with $1=\sum_tw_t$. When users simply want to benchmark their neural compression methodologies on a (sub)set of downstream tasks with known complexity without competing against other teams, the unweighted averaging is the preferred mode of operation for \ifthenelse{\boolean{blind}}{our benchmark}{NeuCo-Bench}. In \cref{sec:cvpranalysis} we report on operational insights related to task weighting as observed in the context of the \ifthenelse{\boolean{blind}}{}{CVPR EARTHVISION }data challenge. The results underline that the concept of \textit{task competitiveness} bears further opportunities for continued research.

\subsection{Platform and Infrastructure}
\label{sec:backend_cvpr}

\begin{figure*}[t]
    \centering
    \includegraphics[width=0.7\linewidth]{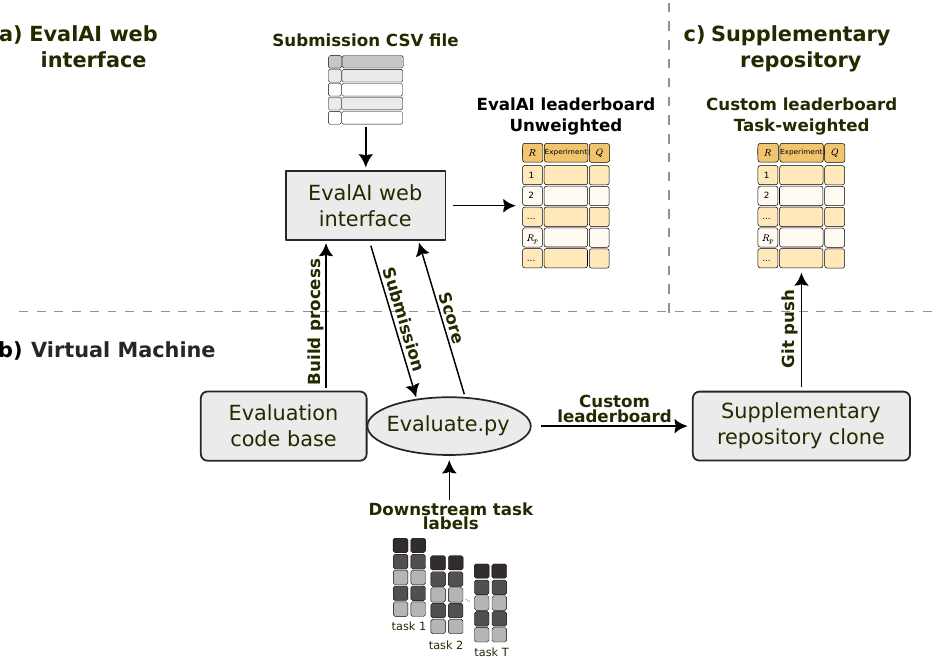}
    \caption{Components and interaction of the data challenge. The community platform Eval.AI (a) interacts with a virtual machine hosted at a cloud service (b). The virtual machine returns the quality score of the submission to the Eval.AI leaderboard and pushes updates to the custom leaderboard (c).}
    \label{fig:Juelich_VM_and_EvalAI_Setup}
\end{figure*}

The core evaluation pipeline was implemented on a virtual machine (VM) with specifications:
\begin{itemize}[nosep, topsep=0pt]
  \item Operating System/OpenStack Image: Ubuntu Jammy 22.04 LTS
  \item CPU: 8 vCPUs, no GPU
  \item RAM: 16GB
  \item Disk: 20GB (OS) + 200GB (data storage)
\end{itemize}
running on top of \ifthenelse{\boolean{blind}}{}{the Juelich Supercomputing Center's (JSC) }OpenStack\footnote{\url{https://www.openstack.org}} cloud environment. The communication with the Eval.AI API for fetching submission data and writing results back to the Eval.AI leaderboard was based on the Eval.AI GitHub \textit{remote challenge evaluation} template\ifthenelse{\boolean{blind}}{}{\footnote{\url{https://github.com/Cloud-CV/EvalAI-Starters/commit/8338085c6335487332f5b57cf7182201b8499aad}}} utilizing the \texttt{requests} Python library.\footnote{\url{https://docs.python-requests.org}} \Cref{fig:Juelich_VM_and_EvalAI_Setup} illustrates the entire setup: (a) the Eval.AI web interface\ifthenelse{\boolean{blind}}{}{\footnote{\url{https://eval.ai/web/challenges/challenge-page/2465}}} and a \textit{supplementary repository}\ifthenelse{\boolean{blind}}{}{\footnote{\url{https://github.com/DLR-MF-DAS/embed2scale-challenge-supplement}}} on one end, and (b) the evaluation procedure which runs on the VM at JSC, on the other end.

As evaluation method (\texttt{Evaluate.py}), \ifthenelse{\boolean{blind}}{our benchmark framework}{NeuCo-Bench\footnote{\url{https://github.com/embed2scale/NeuCo-Bench/tree/040ca567da4d231ced78a16448d2039a0e871276}}} was incorporated into the Eval.AI remote challenge evaluation template running on the VM, cf.\ \textit{Local Repository -- Evaluation Codebase} in \cref{fig:Juelich_VM_and_EvalAI_Setup}. Updates to the Eval.AI challenge web interface got triggered by \textit{GitHub Actions}.\footnote{\url{https://docs.github.com/en/actions}} In addition, the \textit{Supplementary Repository}\ifthenelse{\boolean{blind}}{}{\footnote{\url{https://github.com/DLR-MF-DAS/embed2scale-challenge-supplement}}} serves two purposes:
\begin{itemize}[nosep, topsep=0pt]
    \item for the challenge participants to provide instructions and code examples with options to raise issues, and
    \item to host a \text{Custom Leaderboard} implementing the global ranking introduced in \cref{sec:sup_challenge_eval}, not natively supported by Eval.AI
\end{itemize}
The VM runs a \texttt{cron}job to restart \texttt{Evaluate.py} in case the application terminated. In fact, every minute \ifthenelse{\boolean{blind}}{our benchmark framework}{NeuCo-Bench} polls Eval.AI for new \textit{Submissions} to score. Thereafter, the VM reports $Q$ of the evaluated submission to the \textit{Eval.AI Leaderboard}. It also updates the global \textit{Custom Leaderboard} in the GitHub \textit{Supplementary Repository}.

\subsection{Competition Analysis}
\label{sec:cvpranalysis}

The interaction between participants and organizers through GitHub issues allowed for transparent and traceable communication. In particular, we highlight an update to the challenge that improved comparability between participants by reducing variability in case the same submission is submitted multiple times.\ifthenelse{\boolean{blind}}{}{\footnote{\url{https://github.com/DLR-MF-DAS/embed2scale-challenge-supplement/issues/8}}}

Other learnings from the develploment phase are: 
\begin{itemize}[nosep, topsep=0pt]
    \item Normalization of the target labels across all downstream tasks may be necessary to avoid hyperparameter tuning of the linear probe.
    \item Before normalizing the target labels, the range of the target labels heavily affected the ability of the linear probe to learn a specific task within the given network initialization, learning rate and number of epochs.
\end{itemize}
In total, nine teams participated publicly in the final phase of the \ifthenelse{\boolean{blind}}{}{2025 CVPR EARTHVISION }data challenge, competing over scoring top rank and highest mean $Q$ value across all tasks.

\begin{figure}[htb]
    \centering
    \includegraphics[width=\columnwidth]{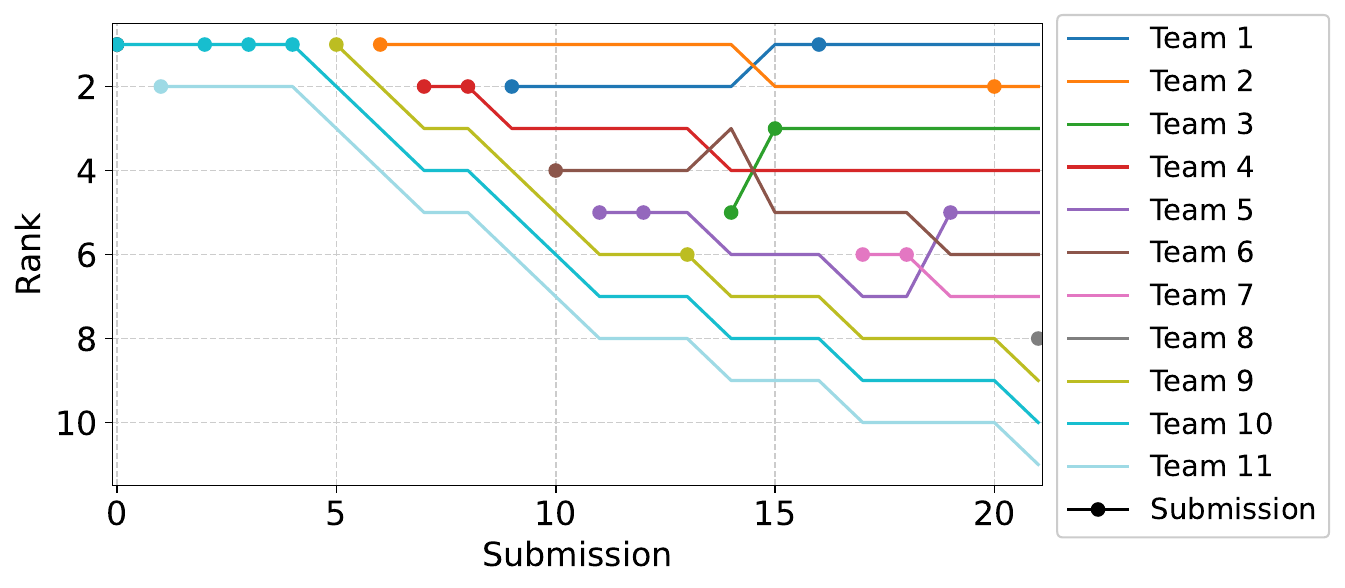}
    \caption{The evolution of participant rankings in the challenge test phase. Lines correspond to participating teams and dots to submissions by the corresponding team. Team 10 is the simple mean baseline described in \cref{sec:results_methods}, and Team 11 is a randomized baseline with randomly sampled, normally distributed embeddings.}
    \label{fig:rank_evolution_test_phase}
\end{figure}

In general, the ranking and the mean $Q$ value are close to identical. However, the team achieving third place upended the order of first and second place, with the effect that the runner-up team achieved a slightly higher mean $Q$-score than the winners. This effect is driven by a change in task weights caused by the third-place team's performance. We note the dynamics around Submissions 14 and 15 as illustrated in \cref{fig:rank_evolution_test_phase} where the ranking dynamics given a sequence of \textit{Submission}s (dots) is documented: Team 1 through 9 are competing numerically indexed by their final position in the challenge ranking. Team 10 represents the simple mean baseline case described in the main paper with additional details in \cref{sec:results_methods}, and Team 11 is a \textit{randomized} baseline submitting randomly sampled, normally distributed embeddings.

\begin{table*}[t]
\centering
\small
\caption{Empirical runtime (in seconds) for different tasks under varying embedding size ($N$), number of epochs ($E$), and number of CV folds ($k$) on single-CPU commodity hardware. Vertical lines separate configurations with different embedding sizes.
}
\label{tab:runtime}
\begin{tabular}{p{3.5cm}|ccccc|cc|cc}
\toprule
\multirow{3}*{\makecell{\textbf{Task} (\# samples)}} &
\multicolumn{5}{c|}{$N=1024$} &
\multicolumn{2}{c|}{$N=512$} &
\multicolumn{2}{c}{$N=2048$} \\
& \makecell{$E=10$\\$k=40$} & \makecell{$E=20$\\$k=20$} & \makecell{$E=20$\\$k=40$} 
& \makecell{$E=40$\\$k=40$} & \makecell{$E=20$\\$k=80$}
& \makecell{$E=20$\\$k=20$} & \makecell{$E=20$\\$k=40$}
& \makecell{$E=20$\\$k=20$} & \makecell{$E=20$\\$k=40$} \\
\midrule
\makecell[l]{Biomass (2415)}            & 4.23 & 4.24 & 7.98 & 16.01 & 16.54 & 3.87 & 7.54 & 4.25 & 8.32 \\
\makecell[l]{Crops (3355)}              & 5.60 & 5.56 & 11.26 & 22.15 & 22.00 & 5.09 & 10.35 & 5.94 & 12.34 \\
\makecell[l]{Clouds (1140)}             & 2.04 & 2.03 & 3.94 & 7.76 & 7.59 & 1.91 & 3.69 & 2.13 & 4.12 \\
\makecell[l]{Landcover\\Agriculture (4691)} & 7.70 & 8.06 & 15.75 & 31.03 & 30.75 & 7.08 & 14.26 & 8.07 & 16.40 \\
\makecell[l]{Landcover Forest (4691)}  & 7.86 & 7.81 & 15.48 & 31.06 & 30.47 & 7.08 & 14.37 & 8.06 & 16.64 \\
\makecell[l]{Heatisland (1659)}         & 2.79 & 2.80 & 5.53 & 10.84 & 10.83 & 2.66 & 5.25 & 2.89 & 5.85 \\
\makecell[l]{No-data (13260)} & 22.45 & 22.45 & 44.30 & 88.68 & 88.18 & 20.46 & 42.03 & 23.18 & 46.60 \\
\bottomrule
\end{tabular}
\end{table*}

A hallmark of our dynamic (global) ranking $R^{(p)}$ can be observed as follows: At submission 14, the submission of Team 3 modified the task weights such that the position of Team 4 and 5 were swapped, even though Team 3 ranked below the two other teams. The same occurred at submission 15, where the submission of team 3 changed the task weights to the benefit of Team 1. The adaptations of the task weights were small compared to the weights of the four tasks with highest weights---on the order of a few percent of the task weights. This drastic effect on the rank of the first two positions is partly due to Team 1 and 2 being neck and neck, Team 1 winning with a weighted average rank 2.31 and Team 2 coming second with 2.44, even though Team 2 scored 15.2 mean $Q$, ahead of Team 1 on 14.9.

The proposed task weighting method as defined by \cref{eq:global_rank_score} achieved to balance the importance of tasks. We noted that the agriculture and forest related tasks were well solved by several teams. The other tasks turned out more challenging. As expected, the (random) baselines were indicated low performers according to $R^{(p)}$. A limitation of the weighting we observed: Since the weighting of the tasks $t$ in \cref{eq:global_rank_score} is based on the variations of the participants for that given task $t$, a participant $p$ very poorly performing by design---such as the random baseline (Team 11)---artificially inflates the task weight when all the other participants perform well. A sensible extension of \ifthenelse{\boolean{blind}}{our benchmark}{the NeuCo-Bench} framework as discussed in \cref{sec:neucobench_framework} will depend on a careful design of downstream tasks and corresponding baselines.

\ifthenelse{\boolean{blind}}{}{Since the competition concluded in early April 2025, the one winner and the fourth-place team have open-sourced their solutions:
the team achieving highest mean $Q$ is available,\footnote{\url{https://github.com/KerekesDavid/embed2scale-solution}} and the fourth-place team, who avoided (pre-)training completely by utilizing MOSAIKS \citep{rolf2021mosaiks} is available,\footnote{\url{https://github.com/isaaccorley/temporal-mosaiks}} too.}

\subsection{Compute Resources \& Tuning Parameters}
\label{sec:compute-and-tuning}

At the \ifthenelse{\boolean{blind}}{}{2025 CVPR EARTHVISION }data challenge,  \ifthenelse{\boolean{blind}}{our benchmark}{NeuCo-Bench} completed evaluations for a single submission within about 10 minutes for the embedding dimension of $N=1024$, the number of epochs per fold are $E=20$, the number of training- and test set splits had been set to $k=40$ in the development phase and $k=200$ in the evaluation phase. The evaluation script ran across a diverse set of eight downstream tasks for real-world geospatial applications. As alluded in \cref{sec:standalone}, users can flexibly adjust evaluation parameters in order to tune the runtime of the standalone implementation\ifthenelse{\boolean{blind}}{}{ of NeuCo-Bench}:
\begin{itemize}[nosep, topsep=0pt]
\item Embedding dimension ($N$, \texttt{embedding\_dim})
\item Number of epochs per CV fold ($E$, \texttt{epochs})
\item Number of CV folds ($k$, \texttt{k\_folds})
\item Choice of tasks included (\texttt{task\_filter})
\end{itemize}
Empirical runtime measurements confirm an approximately linear scaling w.r.t.\ both, the number of epochs $E$ and the number of cross-validation folds $k$. A similar scaling behavior was numerically verified for the dataset size (\# samples) for fixed downstream task. Runtimes are further influenced--—though to lesser extent—--by the embedding dimensionality $N$. For example, increasing the embedding size from 512 to 1024 dimensions results in a runtime increase of approximately 5\% to 10\% across tasks. For $N=1024$ to $N=2048$ dimensions implies an additional increase of about 5\% to 15\%---depending on the task dataset size. Such a sub-linear scaling may be attributed to computation overheads and system-level inefficiencies. Those may reduce the relative computational costs when increasing the embedding dimensionality $N$.
\Cref{tab:runtime} lists a collection of recorded execution times (in seconds) for various parameter configurations per downstream task. All runtimes were measured on a single commodity ARM64 CPU with 16 cores (4.06 GHz) and 64 GB of RAM.

\section{An Extendable Framework}
\label{sec:neucobench_framework}
\begin{table*}[t!]
\caption{\label{tab:NeuCoBenchCompare}Qualitative comparison of our benchmark, PANGAEA, and GEO-Bench}
\centering
\begin{tabular}{llll}
\hline
\textbf{} & \textbf{Ours} & \textbf{PANGAEA} & \textbf{GEO-Bench} \\
\hline
\it Domain & General purpose compression & Geo. foundation models & Geo. foundation models \\
\it Compute & commodity hardware & AI accelerator & AI accelerator \\
\it Model access & Not required & Intermediate features & Backbone finetuning \\
\it Tasks & Classification & Classification & Classification \\
      & Regression & Regression & - \\
      & - & Segmentation \bf(focus) & Segmentation \\
\it Leaderboard API & \tt JSON & \tt JSON & - \\
\hline
\end{tabular}
\end{table*}

Based on our insights from the \ifthenelse{\boolean{blind}}{}{CVPR EARTHVISION 2025 }data challenge, we took our approach to the next level with the intention to build a community around benchmarking compact neural embeddings. \Cref{tab:NeuCoBenchCompare} provides a high-level comparison on how \ifthenelse{\boolean{blind}}{our benchmark}{NeuCo-Bench} fits into existing, popular geospatial benchmarking frameworks. In summary, \ifthenelse{\boolean{blind}}{our benchmark}{\textbf{NeuCo-Bench}} fills the following gaps:
\begin{itemize}
    \item Quantifies the quality of small embeddings based on a variety of downstream tasks without fine-tuning of any neural network backbone.
    \item Provides a standalone toolkit for rapidly benchmarking any compressed embeddings beyond foundation models. In contrast to GEO-Bench and PANGAEA, \ifthenelse{\boolean{blind}}{our benchmark}{the NeuCo-Bench} framework is readily adapted to any compression scenario given:
    \begin{itemize}
        \item Users provide embeddings $z$ where their encoder $E$ takes care of data formats.
        \item Downstream tasks are shared with \ifthenelse{\boolean{blind}}{our benchmark framework}{NeuCo-Bench} as simple CSV files.
    \end{itemize}
    \item Supplies a multi-task performance metric that quantifies embedding size ($N$) vs. downstream accuracy ($Q$).
\end{itemize}

\subsection{Benchmark Tasks}
\label{sec:downstreamtasks}

\begin{figure*}[t]
   \centering
   \includegraphics[width=0.8\textwidth]{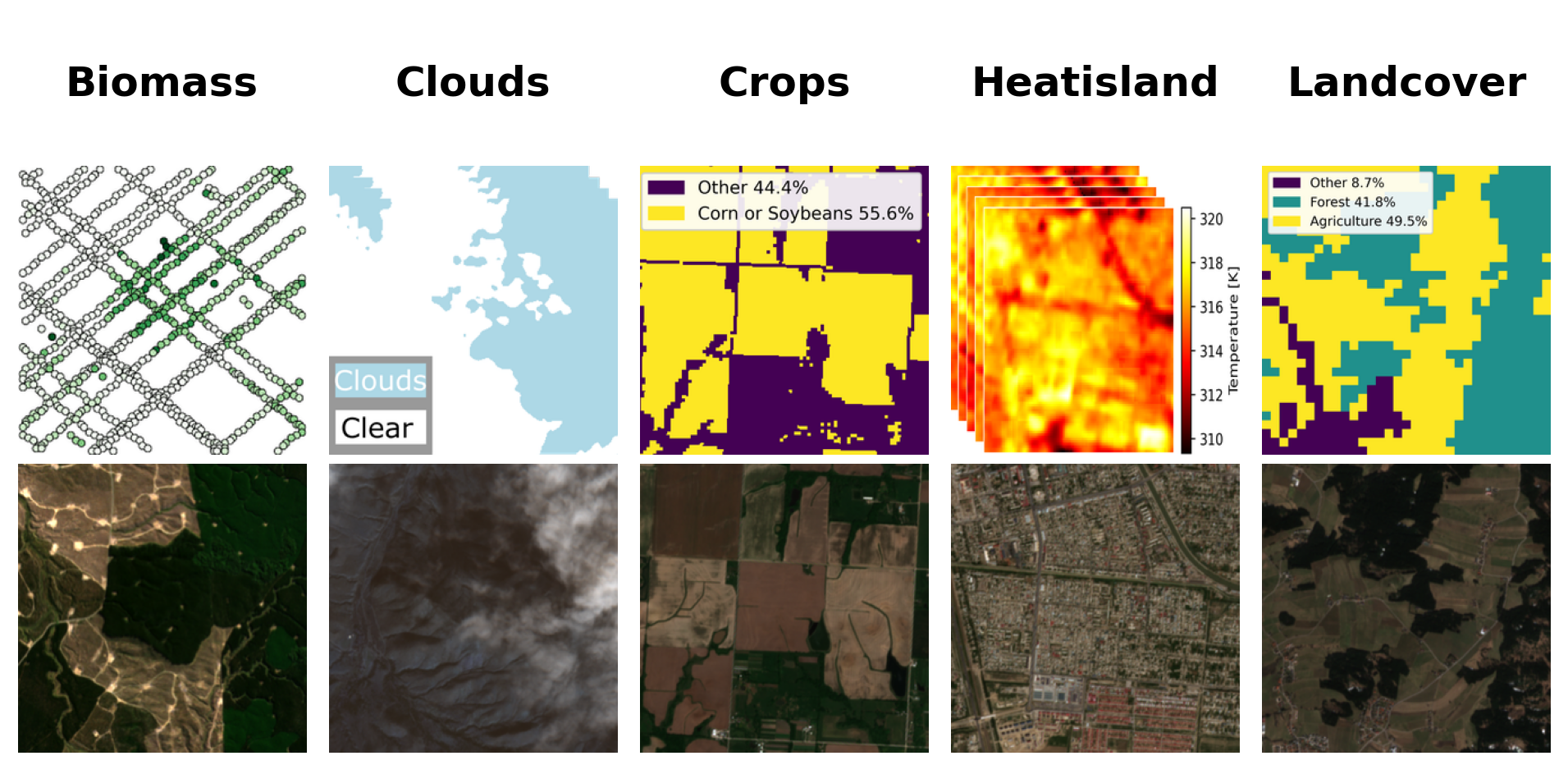}
   \caption{Visualisation of downstream-task labels (top row) and corresponding Sentinel-2 images (bottom row). Although Sentinel-1 is included in every data cube, it was excluded from this visualisation. The cloud and heat island labels are based on aggregated images. }
 \label{fig:overview_downstream}
 \end{figure*}

\Cref{fig:overview_downstream} illustrates examples of Sentinel-2 inputs alongside the corresponding labels.  \Cref{tab:task_descriptions} lists a detailed overview of the derived 11 tasks, whereby 9 of these were used in the \ifthenelse{\boolean{blind}}{}{CVPR EARTHVISION 2025 }data challenge and are highlighted with green check marks. \ifthenelse{\boolean{blind}}{}{Each task is linked to a HuggingFace identifier, following the definition in \cite{SSL4EO-downstream} and listed in \cref{tab:task_descriptions}. }\emph{Clouds} and \emph{Nodata} were not included in the \ifthenelse{\boolean{blind}}{}{CVPR }competition but are provided with the release of \ifthenelse{\boolean{blind}}{our benchmark framework}{NeuCo-Bench}. The task \emph{Random} provides randomly generated labels and associated data cubes from the \emph{Cloud} task, which introduced an additional quality assessment.

The processing pipeline for all presented datasets utilized GEE \cite{GEE} to download data cubes, applying a maximum cloud coverage filter of 10\%, as provided by the GEE property \texttt{CLOUD\_COVER}, except for the clouds task where no restrictions on cloud cover were enforced. Each data cube was aligned to the center of the corresponding label and processed to a size of 264 x 264 pixels. All sample locations with less than 4 images were discarded\ifthenelse{\boolean{blind}}{}{, following the requirements in \cite{blumenstiel2025ssl4eos12v11}}. Whenever possible, only locations that cover all four seasons for Sentinel-1 and Sentinel-2 were chosen. In case of missing latitude and longitude, locations were randomly selected from shapefiles representing regions such as mainland Europe or areas within the US Corn Belt. \ifthenelse{\boolean{blind}}{}{Comparison between the presented dataset and the data provided in \cite{blumenstiel2025ssl4eos12v11}}show one major difference. In January 2022, ESA introduced a new baseline for Sentinel-2 data, effectively shifting all pixel intensities by 1000 units upward. The dataset presented in this work follows the format of GEE, i.e. removing this translation such that the minimum value for Sentinel-2 pixels are 0 both before and after the change by ESA. \ifthenelse{\boolean{blind}}{}{\citet{blumenstiel2025ssl4eos12v11} adheres to the ESA standard and enforces a lower bound for Sentinel-2 pixels of 1000, before and after the change.} To allow for seamless integration between the two datasets, the dataloaders provided in \ifthenelse{\boolean{blind}}{our benchmark}{NeuCo-Bench} includes a setting that toggles a shift by 1000, aligning the distributions of the two datasets.

The \emph{Heat-island} task required additional pre-processing, as Landsat-8 band 10 (B10) was utilized for label generation. This dataset considers only cities with populations exceeding 20,000 and a latitude between 8$^{\circ}$ and 70$^{\circ}$ north. The labels are based on all available Landsat-8 observations from June to the end of August for the years 2021 to 2024 inclusive. In addition, to reduce the impact of remaining clouds, any pixel with a combined brightness (red channel + blue channel + green channel) exceeding 30\% of the maximum possible value or with a B10 temperature lower than 273 K are removed. Images with more than 10\% removed pixels were dropped. The northernmost locations were verified to have average summer temperatures above freezing. For each location, the remaining images are flattened and concatenated over time, and then the mean and standard deviation are calculated from all pixels. The task is to estimate these spatio-temporal statistics. 
\newcommand{\spacedhline}{\addlinespace[0.3em]\hline\addlinespace[0.3em]}
\begin{table*}[t]
  \caption{Descriptions of the downstream tasks provided by the initial release of our benchmark. The tasks used in the
  data challenge are indicated with green check marks in the ``Challenge'' column. The task names correspond to the identifiers as used in the corresponding dataset released.} 
  \label{tab:task_descriptions}
  \centering
  \renewcommand{\arraystretch}{1.0}
  \begin{tabular}{p{2.5cm} p{4cm} p{1.5cm} p{6cm}}
    \toprule
    \textbf{Task} & \textbf{\ifthenelse{\boolean{blind}}{}{HuggingFace }file} & \textbf{Challenge} &\textbf{Description} \\
    \midrule
    
     \textbf{Biomass} & biomass\_mean\_\_regr & \makecell{\textcolor{green}{\checkmark}}& Regression tasks: Biomass density (Mg/ha) mean and standard deviation estimated for pixel-level labels derived from GEDI \citep{GEDI} and sampled based on \citep{Sialelli2025}. \\
     & biomass\_std\_\_regr & \makecell{\textcolor{green}{\checkmark}} & \\
     \spacedhline
     \textbf{Crops} & crops\_\_regr &\makecell{\textcolor{green}{\checkmark}} & Regression task: Combined fraction of Soybean and Corn in the label image \citep{Boryan_CDL}. \\
     \spacedhline
     \textbf{Landcover} & landcover\_agriculture\_\_regr &\makecell{\textcolor{green}{\checkmark}} & Regression task: Percentage of agriculture pixels in the Corine Land cover image \citep{CLC2018}. \\
     & landcover\_forest\_\_regr &\makecell{\textcolor{green}{\checkmark}} & Regression task: Percentage of forest pixels in the Corine Land cover image. \\
     \spacedhline
     \textbf{Clouds} & clouds\_reg\_\_regr & \makecell{\textcolor{red}{\XSolid}} & Regression task: Average cloud cover fraction across four seasons in one year \citep{Ayabar2024CloudSen12}. \\
     \spacedhline
     \textbf{Heatisland} & heatisland\_mean\_\_regr  & \makecell{\textcolor{green}{\checkmark}} &
     Regression tasks: Summer surface temperature mean and standard deviation in Kelvin based on LandSat-8 \citep{landsat8}.\\
     & heatisland\_std\_\_regr & \makecell{\textcolor{green}{\checkmark}} & \\
     \spacedhline
     \textbf{Nodata} & nodata\_\_regr &\makecell{\textcolor{red}{\XSolid}} & Regression task: Fraction of pixels with value zero in a Sentinel-2 image (based on all 13,260 available samples). \\
     
     \textbf{Random} & random\_reg\_\_regr &\makecell{\textcolor{green}{\checkmark}} & Regression task: Random task with a majority of zero labels. The data cubes are the same as for Clouds. \\
     & random\_cls\_\_cls &\makecell{\textcolor{green}{\checkmark}} & Classification task: Random binary classification for a majority of zero labels. The data cubes are the same as for Clouds.\\
    \bottomrule
  \end{tabular}
\end{table*}

\textbf{Public vs.\ Secret Downstream Tasks.} We released the hidden \ifthenelse{\boolean{blind}}{}{NeuCo-Bench }downstream tasks (cf. \cref{tab:task_descriptions} with green check mark) after the conclusion of the \ifthenelse{\boolean{blind}}{}{2025 CVPR EARTHVISION }workshop to make publicly accessible the standalone \ifthenelse{\boolean{blind}}{}{SSL4EO-S12-}downstream dataset for reasons of transparency, and to be used and contributed to by the neural compression community. As common with public benchmarks designed for standalone usage, we assume that \ifthenelse{\boolean{blind}}{benchmark}{NeuCo-Bench} users would not jeopardize the developments of their own compressor $E$ by willingfully exploiting knowledge of the downstream tasks they test on. Removing and adding (mix-and-match) downstream tasks \ifthenelse{\boolean{blind}}{}{in NeuCo-Bench }for a new competition avoids overfitting of a state-of-the-art compressor $E$. The process is as straightforward as uploading such data $x$ to \ifthenelse{\boolean{blind}}{a file sharing service}{HuggingFace}, i.e.,
\begin{enumerate}
    \item \textit{public}: Each data point $x_i$ just needs a unique (identified by hash $i$) name (cf. e.g., directory \texttt{data/} of \ifthenelse{\boolean{blind}}{}{SSL4EO-S12-}downstream dataset) to upload corresponding \dots
    \item \textit{secret till conclusion of competition}: \dots label CSV files (cf. e.g., directory \texttt{labels/} of \ifthenelse{\boolean{blind}}{}{SSL4EO-S12-}downstream dataset, \texttt{id} column of CSV file) \dots
\end{enumerate}
\dots for the \ifthenelse{\boolean{blind}}{benchmark}{NeuCo-Bench} engine to perform its downstream task evaluations.

\subsection{Standalone Python Implementation}
\label{sec:standalone}

In order for a clean separation of code from the open-source platform Eval.AI, we developed a minimal viable standalone Python code base\ifthenelse{\boolean{blind}}{}{\footnote{\url{https://github.com/embed2scale/NeuCo-Bench/tree/040ca567da4d231ced78a16448d2039a0e871276}}} to serve as plug-and-play for any larger ecosystem integrating of the core \ifthenelse{\boolean{blind}}{}{NeuCo-Bench }framework. In fact, as \cref{fig:Juelich_VM_and_EvalAI_Setup} demonstrates, the scoring for the Eval.AI leaderboard is entirely taken care of by \ifthenelse{\boolean{blind}}{our benchmark}{NeuCo-Bench}. Correspondingly, our framework commits an additional, customized leaderboard that \textit{globally} depends on all submissions to a dedicated GitHub repository.\ifthenelse{\boolean{blind}}{}{\footnote{\url{https://github.com/DLR-MF-DAS/embed2scale-challenge-supplement/tree/272c911e8fb527d265de9be18fc81e16208ca0b6?tab=readme-ov-file\#leaderboard}}}

The \ifthenelse{\boolean{blind}}{benchmark}{NeuCo-Bench} core \texttt{evaluation.py} functionality\ifthenelse{\boolean{blind}}{}{\footnote{\url{https://github.com/embed2scale/NeuCo-Bench/blob/040ca567da4d231ced78a16448d2039a0e871276/benchmark/evaluation/evaluation.py}}} separately fetches
\begin{itemize}[nosep, topsep=0pt]
    \item the user's embeddings (submission), \texttt{/path/to/submission\_file.csv}, and
    \item the downstream task annotation data (labels), \texttt{/path/to/annotation\_directory/}
\end{itemize}
as  ASCII-formatted \texttt{CSV} files given predefined local paths and directories\ifthenelse{\boolean{blind}}{}{\footnote{\url{https://github.com/embed2scale/NeuCo-Bench/blob/040ca567da4d231ced78a16448d2039a0e871276/benchmark/main.py\#L14-L19}}} as simple interface entirely independent of Eval.AI. Given any ranking procedure implemented,\ifthenelse{\boolean{blind}}{}{\footnote{\url{https://github.com/embed2scale/NeuCo-Bench/blob/040ca567da4d231ced78a16448d2039a0e871276/benchmark/evaluation/results.py\#L43-L80}}} the resulting leaderboard is saved as human-readable \texttt{JSON} file\ifthenelse{\boolean{blind}}{}{\footnote{\url{https://github.com/embed2scale/NeuCo-Bench/blob/040ca567da4d231ced78a16448d2039a0e871276/benchmark/evaluation/results.py\#L8-L22}}} in a corresponding \texttt{/path/to/results\_directory/}. For downstream (binary) classification tasks, the confusion matrix and related scores such as precision, recall, F1, and overall accuracy are calculated along with the ROC-AUC-score (area under Receiver-Operator-Characteristic graph). For regression, the R-squared, mean squared, and mean absolute errors are computed.\ifthenelse{\boolean{blind}}{}{\footnote{\url{https://github.com/embed2scale/NeuCo-Bench/blob/040ca567da4d231ced78a16448d2039a0e871276/benchmark/evaluation/metrics.py}}}

To serve as seed towards an open-source and open science community, we designed the standalone Python implementation of \ifthenelse{\boolean{blind}}{our benchmark}{NeuCo-Bench} modular for easy extension. Depending on compute resources, we encourage future contributions to add novel probing models, cross validation schemes, and performance scores (cf.\ \cref{eq:quality_score}) beyond the current.\ifthenelse{\boolean{blind}}{}{\footnote{\url{https://github.com/embed2scale/NeuCo-Bench/blob/040ca567da4d231ced78a16448d2039a0e871276/benchmark/evaluation/linear_probing.py}}} As a bonus, our standalone implementation allows to store plots of loss curves, linear correlation of regression tasks, and a confusion matrix for classification on disk.\ifthenelse{\boolean{blind}}{}{\footnote{\url{https://github.com/embed2scale/NeuCo-Bench/blob/040ca567da4d231ced78a16448d2039a0e871276/benchmark/evaluation/visualisations.py}}}

Running \ifthenelse{\boolean{blind}}{our benchmark}{NeuCo-Bench} standalone on the command line reduces to something as simple as:
\begin{lstlisting}[style=BashStyle,basicstyle=\relsize{-2}\ttfamily]
python main.py \
  --annotation_path /path/to/annotation_directory/ \
  --submission_file /path/to/submission_file.csv \
  --output_dir /path/to/results_directory/ \
  --config /path/to/config.yaml \
  --method_name 'method-name' \
  --phase 'phase-name'
\end{lstlisting}
where \texttt{method-name} and \texttt{phase-name} are free strings to define an output (sub-)directory {\tt phase-name/method-name\_YYYYMMDD\_HHmmSS} under {\tt /path/to/results\_directory/},
 with \texttt{YYYYMMDD} a date of year \texttt{YYYY} and zero-padded numerical month \texttt{MM} and day \texttt{DD}. \texttt{HHmmSS} indicates a time of the day in hours \texttt{HH}, minutes \texttt{mm}, and seconds \texttt{SS}, accordingly. A \texttt{YAML} file \texttt{/path/to/config.yaml} specifies details of the evaluation such as:
\begin{lstlisting}[style=PythonStyle,basicstyle=\relsize{-2}\ttfamily]
# number of embedding dimensions
embedding_dim: 1024          
# batch size for (linear) probing
batch_size: 64               
# number of epochs to optimize the (linear) probe for
epochs: 20                   
# learning rate to optimize with
learning_rate: 0.001         
# number of splits to generate statistics over        
k_folds: 40                  
# standardize embeddings by their global mean and std
standardize_embeddings: true 
# normalize in range [0,1]
normalize_labels: true       
# Filter to include only specific tasks.
# all in /path/to/annotation_directory/ per default
# example: ["biomass_mean", "biomass_std"]
task_filter: false           
                             
# etc.
\end{lstlisting}

\subsection{Licenses for Data \& Software}
\label{sec:software_licenses}

\begin{table*}[t]
  \caption{List of licenses related to datasets currently included in our benchmark. All of these, except \textit{Clouds}, are available in GEE. 
  the main paper lists years of target labels, ranging from 2018 through 2024.}
  \label{tab:dataset-licenses}
  \centering
  \renewcommand{\arraystretch}{1.0}
  \begin{tabular}{l p{7.7cm} p{2.7cm}}
    \toprule
    \textbf{Dataset} & \textbf{Origin of Data} & \textbf{License} \\
    \midrule
    Sentinel-1 \& -2  & ESA / Copernicus & CC BY-SA 3.0 IGO \\
    Landsat-8   & USGS \citep{landsat8} & Public Domain \\
    CDL       & USDA NASS Cropland Data Layers \citep{Boryan_CDL} & Public Domain \\
    CORINE   & European Environment Agency (EEA), European Union & Full, Open, \\ 
                & Copernicus Land Monitoring Service \citep{CLC2018} & and Free Access \\
    CloudSen12+ & CloudSEN12 project \citep{Ayabar2024CloudSen12} & CC0 1.0 \\
    GEDI     & NASA \citep{GEDI} & Public Domain \\
    \bottomrule
  \end{tabular}
\end{table*}

\ifthenelse{\boolean{blind}}{Our benchmark}{NeuCo-Bench} builds on open-source software and is released under the Apache 2.0 license\ifthenelse{\boolean{blind}}{}{ publicly available at \url{https://github.com/embed2scale/NeuCo-Bench}}. All package dependencies are listed in the \texttt{requirements.txt}\ifthenelse{\boolean{blind}}{}{\footnote{\url{https://github.com/embed2scale/NeuCo-Bench/blob/main/requirements.txt}}} file, and those are licensed under widely-accepted open-source terms\footnote{\url{https://opensource.org/licenses}}, including BSD, MIT, PSFL, The Unlicense\footnote{code-equivalent to CC0 data licenses}, MPL-2.0\footnote{weak copy-left that allows for integration with non-copyleft licenses}, and Apache. These permissive licenses allow academic research and commercial use, making them fully compatible with the chosen Apache 2.0 license. \Cref{tab:dataset-licenses} lists all data currently included in \ifthenelse{\boolean{blind}}{our benchmark}{NeuCo-Bench} along with their origin. Google Earth Engine (GEE) \citep{GEE} was utilized as the primary platform for downloading downstream task data as introduced in \cref{sec:downstreamtasks}. Future data and code contributions to \ifthenelse{\boolean{blind}}{our benchmark}{NeuCo-Bench} are required to be licensed under CC-BY 4.0 and Apache 2.0, respectively.

We note that the current \ifthenelse{\boolean{blind}}{}{NeuCo-Bench }implementation\ifthenelse{\boolean{blind}}{ of our benchmark}{} lists CUDA packages covered by a proprietary NVIDIA license\footnote{\url{https://docs.nvidia.com/cuda/eula/index.html}}. However, we do neither bundle nor redistributes corresponding binaries. Users and contributors to \ifthenelse{\boolean{blind}}{our benchmark}{NeuCo-Bench} that share related docker containers need to explicitly attribute NVIDIA's license. Fortunately, and as alluded in \cref{tab:runtime} and \cref{sec:backend_cvpr}, \ifthenelse{\boolean{blind}}{our benchmark}{NeuCo-Bench} runs swiftly in a VM with commodity hardware specifications on CPU compute, only. Accordingly, the standalone \ifthenelse{\boolean{blind}}{}{NeuCo-Bench }implementation introduced in \cref{sec:standalone} can be started with (Bash) environment variable \texttt{CUDA\_VISIBLE\_DEVICES=''} to avoid usage of GPU resources.

\section{Baseline Methods}
\label{sec:baseline_methods_eval}
This appendix expands on the general evaluations introduced in 
the main paper by providing methodological details and extended results. Unlike the \ifthenelse{\boolean{blind}}{}{CVPR }challenge setting, which separated development and evaluation splits, all results in 
the main paper and this appendix are computed on the full downstream datasets. Unless noted otherwise, we follow the evaluation protocol introduced in the main text, and use $E=20$ training epochs, $k=50$ train–test splits, and a learning rate of $10^{-3}$. We report raw $R^2$ values, clipping negative scores to $[0,1]$ only for visualization.

We begin by comparing the temporal aggregation methods in \cref{sec:temporal_analysis} which motivate the use of post-encoding aggregation for all subsequent analyses. Thereafter, we report additional details of the baseline methods and results for the 1,024-dimensional embedding setup \cref{sec:results_methods}. In \cref{sec:ablations_and_ext} we extend the ablations introduced in 
the main paper by providing per-task results for varying embedding dimensions and decoder choices beyond linear probing.

\subsection{Temporal Aggregation Analysis}
\label{sec:temporal_analysis}

\begin{figure*}[t]
  \centering
  \includegraphics[width=0.9\textwidth]{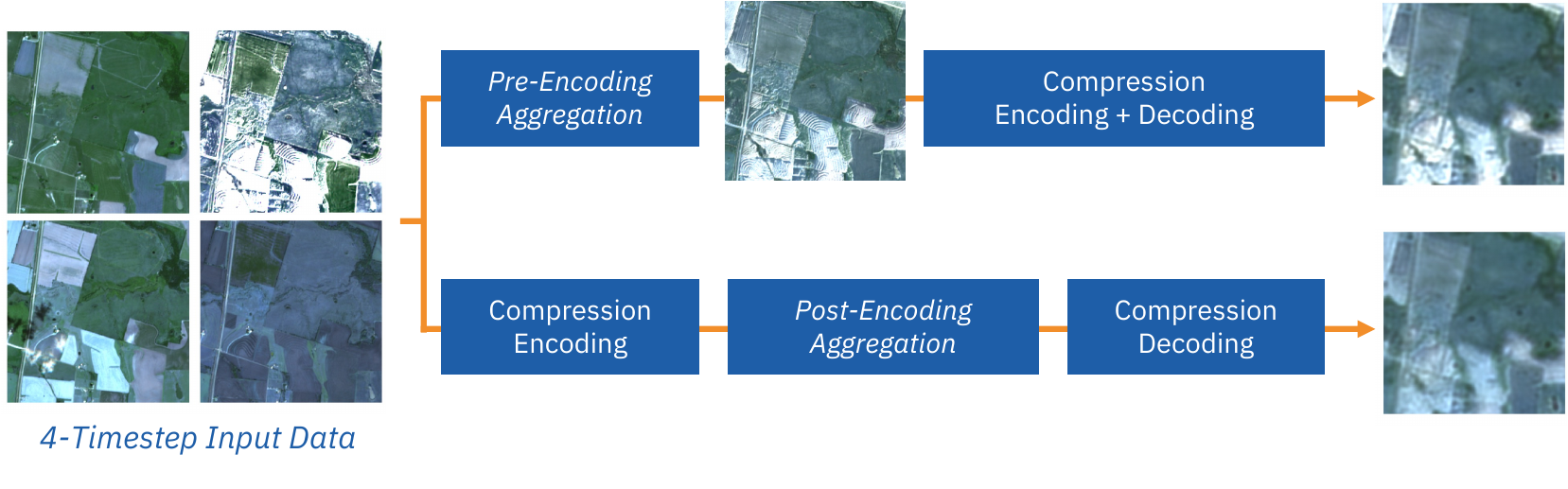}
  \caption{Illustration of pre-encoding vs.\ post-encoding aggregation. In post-encoding, each seasonal image is encoded separately before combining embeddings, which mitigates outlier effects (e.g., snow) but increases runtime fourfold.}
  \label{fig:temporal_aggregation}
\end{figure*}

We study image-based encoding methods for our baseline evaluations. As each input sample contains four seasonal Sentinel-1/2 snapshots, we handle temporal sequence data by reducing the four snapshots into a single embedding using two aggregation strategies:

\begin{itemize}
    \item \textbf{Pre-encoding:} seasonal snapshots are averaged before encoding.
    \item \textbf{Post-encoding:} each snapshot is encoded separately and the resulting embeddings are averaged. As shown in \cref{fig:temporal_aggregation}, this approach better handles seasonal outliers (e.g., snow) at the cost of additional computation.  
\end{itemize}
%

Across all tested methods, post-encoding aggregation provides consistent $R^2$ improvements for most methods and tasks as summarized in \Cref{tab:temporal_summary}. The largest gains occur for the cloud-fraction prediction task, with increases of up to $+0.42$ for ViT-based encoders and $+0.30$ for TerraMind. Semantic tasks such as land-cover show smaller but consistent improvements (+0.011 to +0.016). In summary, post-encoding yields performance gains, in particular for temporally sensitive tasks such as determination of cloud fractions. These results motivate the use of post-encoding aggregation for the baseline results.
\begin{table*}[h]
\centering
\relsize{-.5}
\caption{Comparison of temporal aggregation methods. We show average $\bar R_p^2$ and $\bar R_P^2$ scores across all downstream tasks (see main paper for details) for pre-encoding vs.\ post-encoding aggregation, respectively. We also provide the overall improvement $\boldsymbol{\Delta R^2}=\bar R_P^2 - \bar R_p^2$ and the best gain per task, \textbf{Best~$\boldsymbol{\Delta R^2}$}.}
\label{tab:temporal_summary}

\begin{tabular}{%
>{\raggedright\arraybackslash}m{0.16\textwidth}
>{\centering\arraybackslash}m{0.12\textwidth}
>{\centering\arraybackslash}m{0.12\textwidth}
>{\centering\arraybackslash}m{0.12\textwidth}
>{\centering\arraybackslash}m{0.24\textwidth}
}
\toprule
\textbf{Method} &
\makecell{\textbf{Pre-Enc.}\\$\bar R_p^2$} &
\makecell{\textbf{Post-Enc.}\\$\bar R_P^2$} &
\makecell{$\boldsymbol{\Delta R^2}$} &
\makecell{\textbf{Best} $\boldsymbol{\Delta R^2}$\\\textbf{(task)}} \\
\midrule
\makecell[l]{Averaging\\Baseline}   & -0.522 & -0.385 & +0.137 & +0.270 (clouds) \\
\makecell[l]{Factorized\\Prior}     &  0.238 &  0.233 & -0.005 & +0.044 (clouds) \\
\makecell[l]{DINO\\ResNet}     &  0.289 &  0.397 & +0.108 & +0.318 (clouds) \\
\makecell[l]{DINO ViT}        &  0.382 &  0.470 & +0.088 & +0.423 (clouds) \\
\makecell[l]{MAE ViT}         &  0.481 &  0.537 & +0.056 & +0.272 (clouds) \\
\makecell[l]{TerraMind}       &  0.600 &  0.659 & +0.059 & +0.297 (clouds) \\
\bottomrule
\end{tabular}
\end{table*}

\subsection{Embedding Method Evaluations}
\label{sec:results_methods}
The following section extends the baseline evaluations in the main paper and provides additional details on the tested embedding methods.

\begin{figure*}[ht]
    \centering
    \includegraphics[width=.75\textwidth]{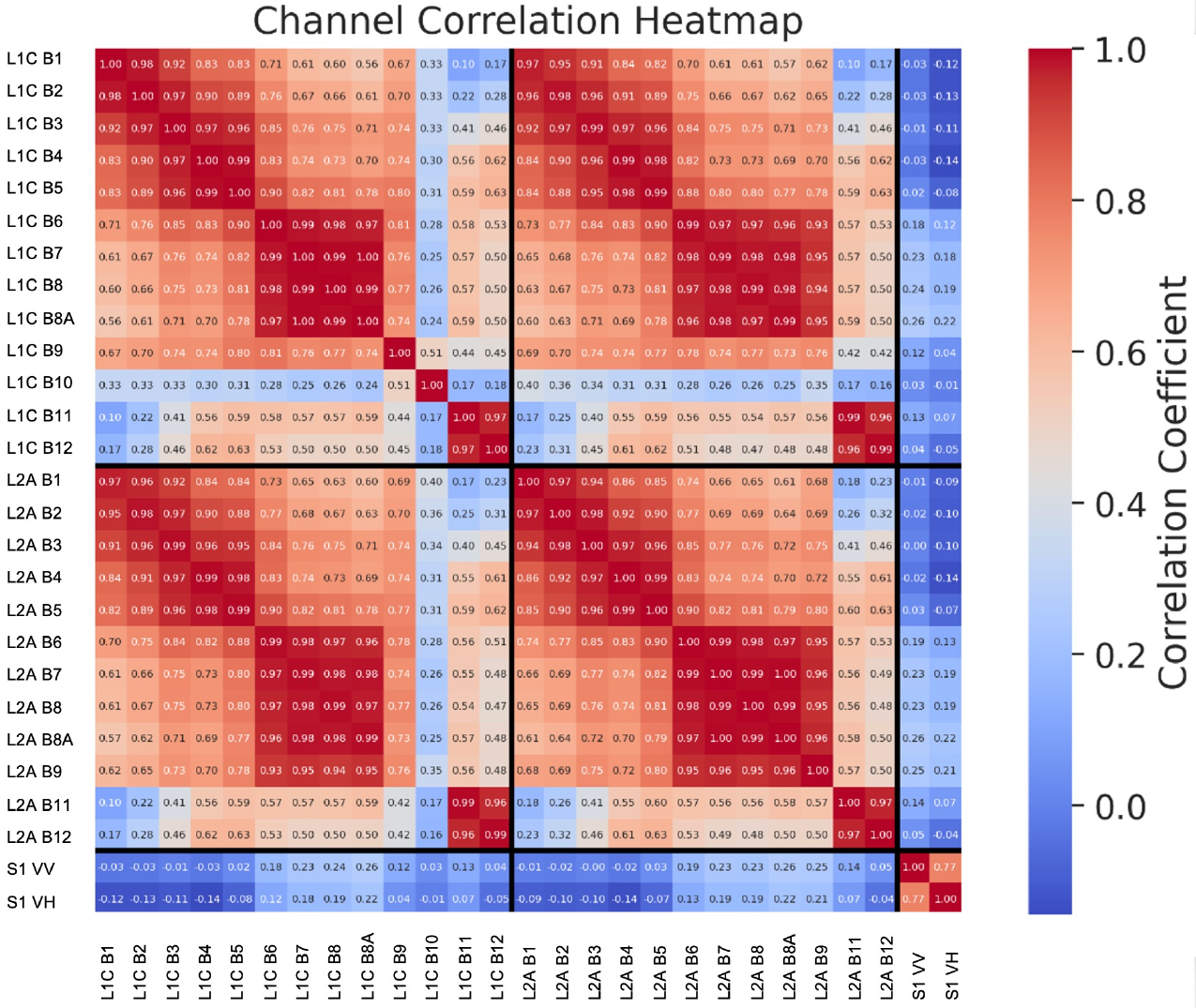}
    \caption{Pearson correlation coefficient matrix of the 27 data cube channels. In here, we abbreviate the channels of the Sentinel-2 L1C and L2A products as L1C and L2A, respectively appending the channel name (B1, B2, \dots).}
    \label{fig:dataset_correlation}
\end{figure*}

\textbf{Averaging Baseline.} As a simple informative reference, we construct a \emph{Mean baseline} by strongly downsampling and averaging the \ifthenelse{\boolean{blind}}{}{SSL4EO-S12 }data cubes. First, we reduce the spatial resolution of each of the 27 channels from 264x264 pixels to 8x8 by bi-linear interpolation. Next, we exploit correlation as visualized by \cref{fig:dataset_correlation} reducing the number of channels from 27 down to four. We average the channels B1 through B9 of both S2L1C and S2L2A and we do similar for channels B11 and B12, respectively. Channel B10 of S2L1C is kept separate since no corresponding band exists in S2L2A.
Seasonal snapshots are kept separate, yielding 8$\times$8$\times$4$\times$4 values, flattened to $N=1024$. This baseline indicates how much task-relevant information survives coarse spatial and spectral aggregation.

\textbf{Neural rate–distortion compressors.}  

\begin{figure}[ht]
  \centering
  \includegraphics[width=0.7\linewidth]{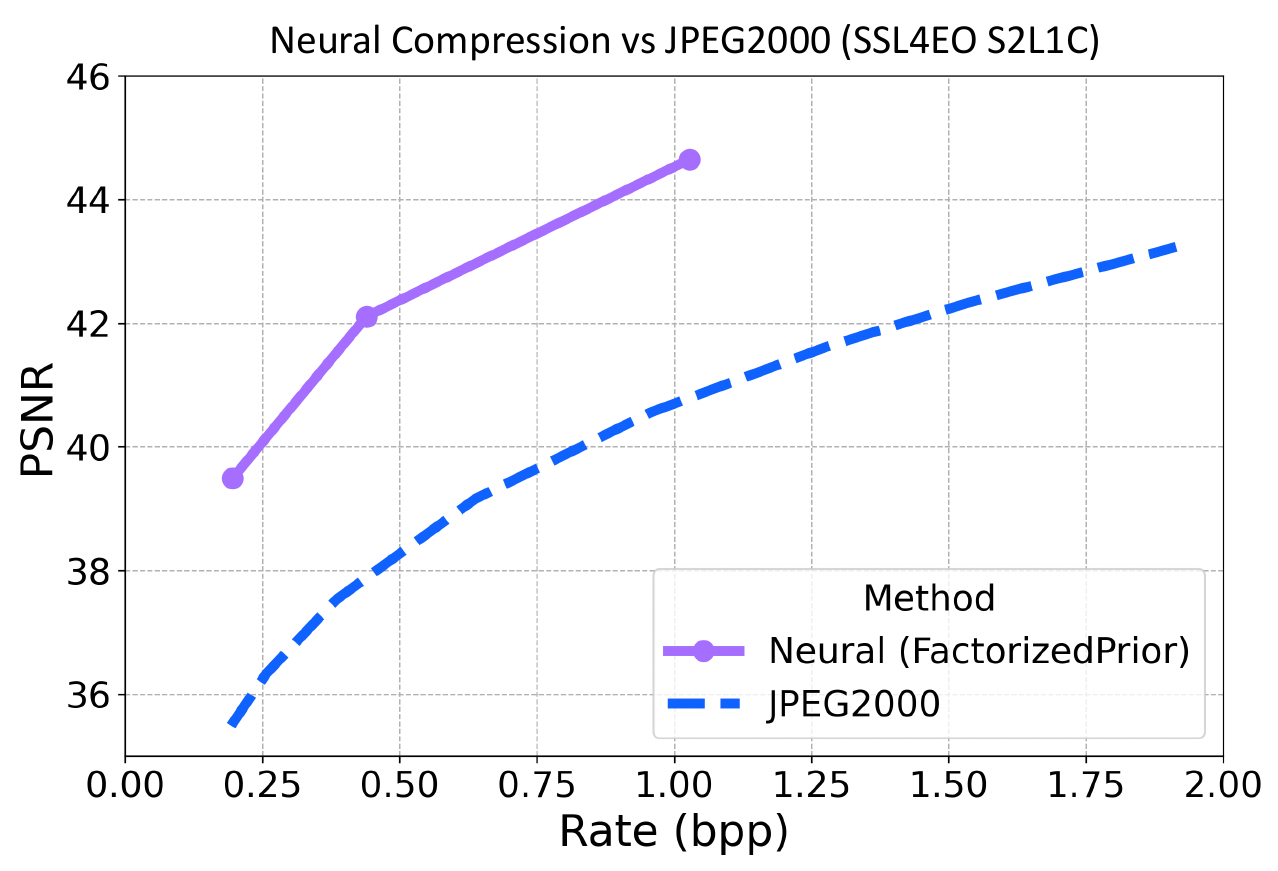}
  \caption{Rate–distortion performance of the Factorized Prior neural compressor, demonstrating superior compression quality over the JPEG2000 baseline.}
  \label{fig:neural_rate_distortion_psnr}
\end{figure}

We adopt and train a Factorized Prior autoencoder~\citet{balleEndtoendOptimizedImage2016} for Sentinel-2 L1C imagery. Models use 256 intermediate channels and 128 latent channels, trained with loss
\begin{equation}\label{eq:rate-distortion-loss}
    \mathcal{L} = R + \lambda D
\end{equation}
where $D$ is MSE distortion and $\lambda \in \{0.025,0.1,0.5\}$ and $R$ represents the entropy-coding term for bit-stream compression. These compressors exceed JPEG2000 PSNR at roughly half the bitrate (Fig.~\ref{fig:neural_rate_distortion_psnr}). At inference, we pool latents to $4\times4$, flatten to 2048, and average adjacent channels to yield 1024-dim embeddings. We evaluate three rate–distortion settings, which control the weight of the entropy loss term during training, cf.\ \cref{eq:rate-distortion-loss}. We observe that the model with the strongest emphasis on compression ($\lambda=0.025$) performs best overall, while lower compression focus ($\lambda=0.5$) degrades performance. However, all variants score behind FM embeddings and struggle with the strong spatial–temporal aggregation and linear-probing setup.

\textbf{Self-supervised foundation models (FMs).}  
We evaluate a broad set of pretrained EO FMs. As introduced in 
the main paper, we apply a consistent embedding aggregation pipeline across all FMs. As motivated by \cref{sec:temporal_analysis}, we utilize temporal post-encoding aggregation throughout our experiments. CNN outputs are reduced via global average pooling. If the pooled feature dimension exceeds 1,024, we apply pairwise channel means, i.e., adjacent feature channels are averaged in pairs to halve the dimensionality while preserving coarse channel structure. For ViT encoders, we average the spatial patch tokens (excluding the CLS token, if present) to form a single embedding. Additionally, we also evaluate CLS-token embeddings for Prithvi and Clay. All aggregated embeddings are padded to the target 1,024-dimensional space. Overall, we observe the following trends:

\begin{itemize}
    \item \textbf{FM Embeddings.} Multimodal models, most prominently TerraMind, consistently outperforms all other embeddings, achieving the highest $R^2$ across most tasks. We highlight that jointly modelling Sentinel-1/2 can provide task benefits under strong spatial and spectral aggregation. DOFA, while scoring below TerraMind, still achieves consistent performance across all task.
    EO-specific, single-modal ViTs such as Prithvi and Clay follow below TerraMind and offer strong results across semantic and geophysical tasks. SSL4EO-pretrained models exhibit complementary strengths, that is: CNN variants perform strongly on land-cover tasks, but fall behind ViTs on geophysical regressions. Notably, S12-DINO ResNet scores the second highest on both the land-cover agriculture and forest tasks. MAE ViT achieves balanced and high performance across tasks, while contrastive DINO and MoCo excel on semantic land-cover tasks. However, DINO\slash MoCo are less competitive on geophysical regression.

    \item \textbf{CLS Token Embeddings.} The comparison between patch-averaged and CLS-token embeddings for Prithvi and Clay demonstrates that mean patch-token averaging is the more robust strategy: for Prithvi, CLS performance is slightly lower but remains close, whereas for Clay the CLS token underperforms patch averaging. 
\end{itemize}

\subsection{Non-linear Probing \& Embedding Size Ablations}\label{sec:ablations_and_ext}
\paragraph{Linear vs.\ Non-Linear Probing.}  
As part of our evaluation design, we explored the impact of decoder complexity on downstream task performance. While linear probing is the default protocol, we deliberately investigated non-linear alternatives to assess whether additional decoder capacity meaningfully improves results.  

\begin{figure*}[t]
  \centering
  \includegraphics[width=.95\textwidth]{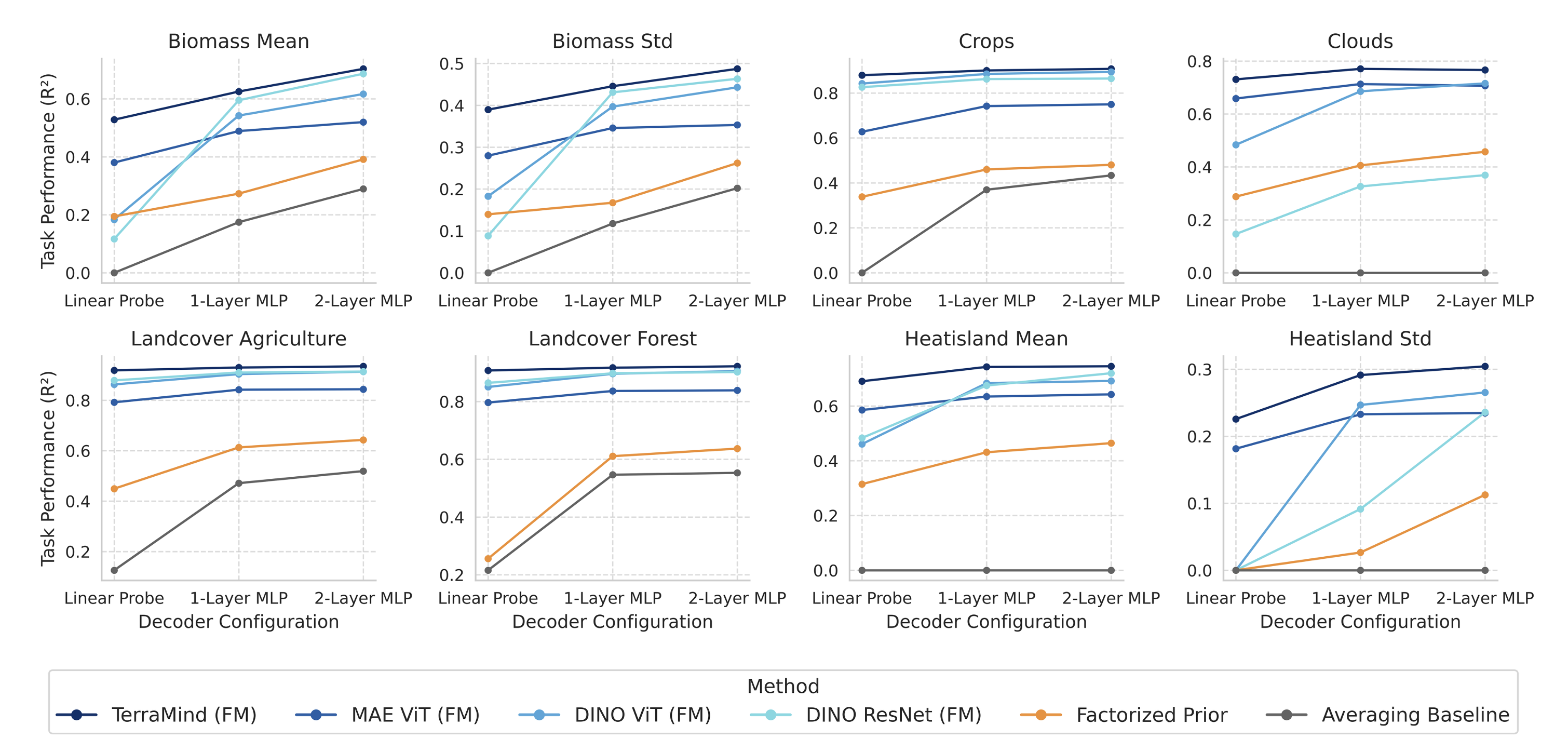}
  \caption{Per-task results for linear vs.\ non-linear probes. Non-linear decoders benefit weaker embeddings but have little effect on top-performing methods.}
  \label{fig:indiv_decoder_analyses}
\end{figure*}

Additionally to the main paper plot on average task scores, we provide per task plots in Figures~\ref{fig:indiv_decoder_analyses} comparing linear probes with one- and two-hidden-layer MLP decoders. We observe three consistent findings:  
\begin{itemize}
    \item \textbf{Stable rankings.} The relative ranking of embedding methods remains nearly the same across probe types, indicating that differences between methods are not an artifact of probe capacity.  
    \item \textbf{Marginal gains for strong embeddings.} Top-performing embeddings (e.g., TerraMind, MAE) improve by less than 0.06 $R^2$ on average when switching to non-linear probes, demonstrating that these embeddings are already highly linearly expressive.  
    \item \textbf{High computational overhead.} Increasing decoder depth leads to $\sim$170$\times$ and $\sim$464$\times$ more parameters for one and two hidden layers, respectively, with only small performance gains.  
\end{itemize}

Interestingly, weaker embeddings benefit disproportionately from non-linear probes, suggesting that added decoder complexity can compensate for lower-quality representations. However, this comes at substantial computational cost.  

Taken together, these results highlight that linear probing is not only efficient but also a discriminative evaluation strategy: it faithfully reflects the intrinsic quality of embeddings while enabling scalable benchmarking. Non-linear decoders may be useful for future extensions to more complex tasks (e.g., pixel-wise segmentation), but for the image-level tasks studied here, linear probing provides a robust and interpretable measure of embedding quality.  

\begin{figure*}[t]
  \centering
  \includegraphics[width=.94\textwidth]{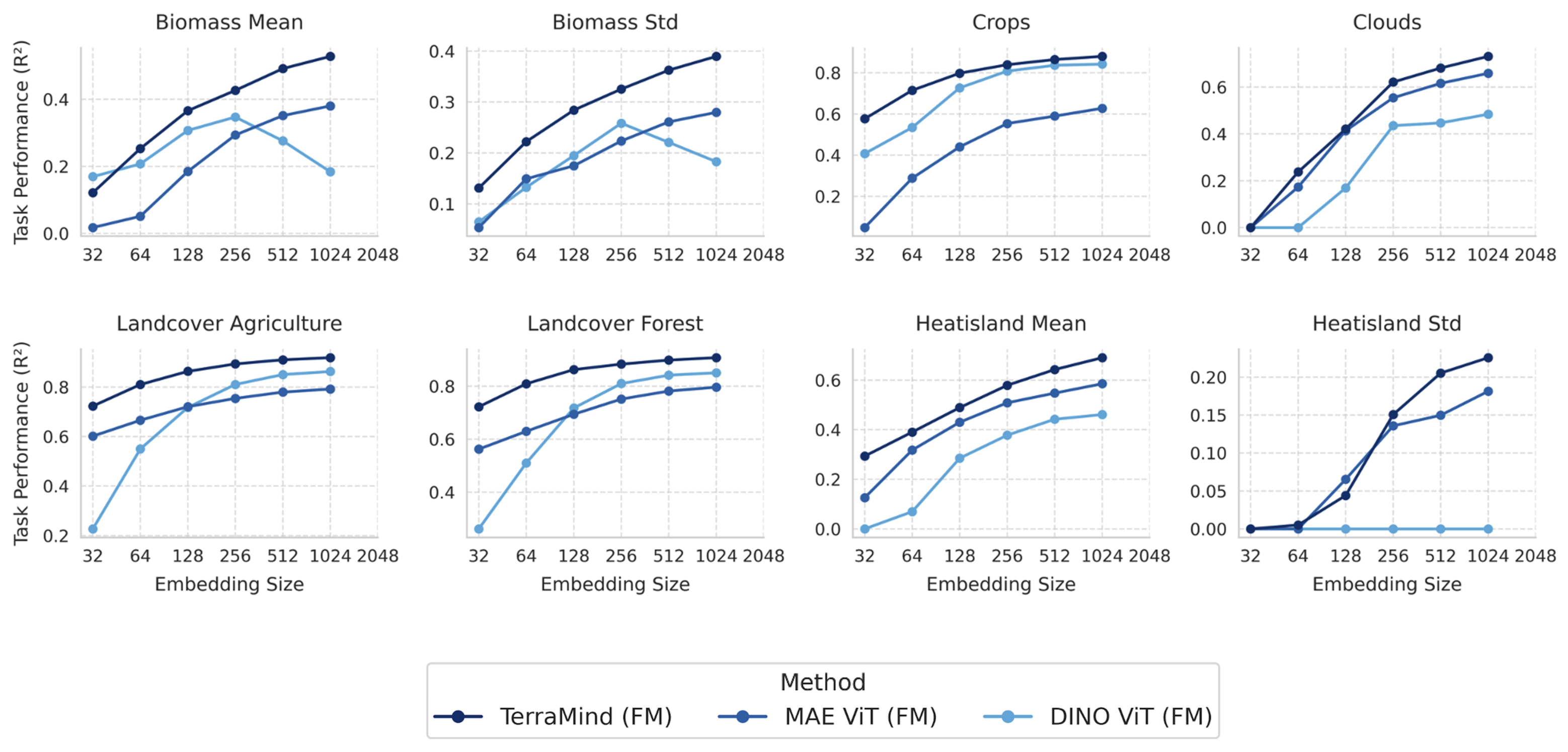}
  \caption{Embedding size ablation for ViT-based models. Performance peaks at the native patch embedding size and drops with reduced dimensions.}
  \label{fig:size_vit}
\end{figure*}
\begin{figure*}[t]
  \centering
  \includegraphics[width=.94\textwidth]{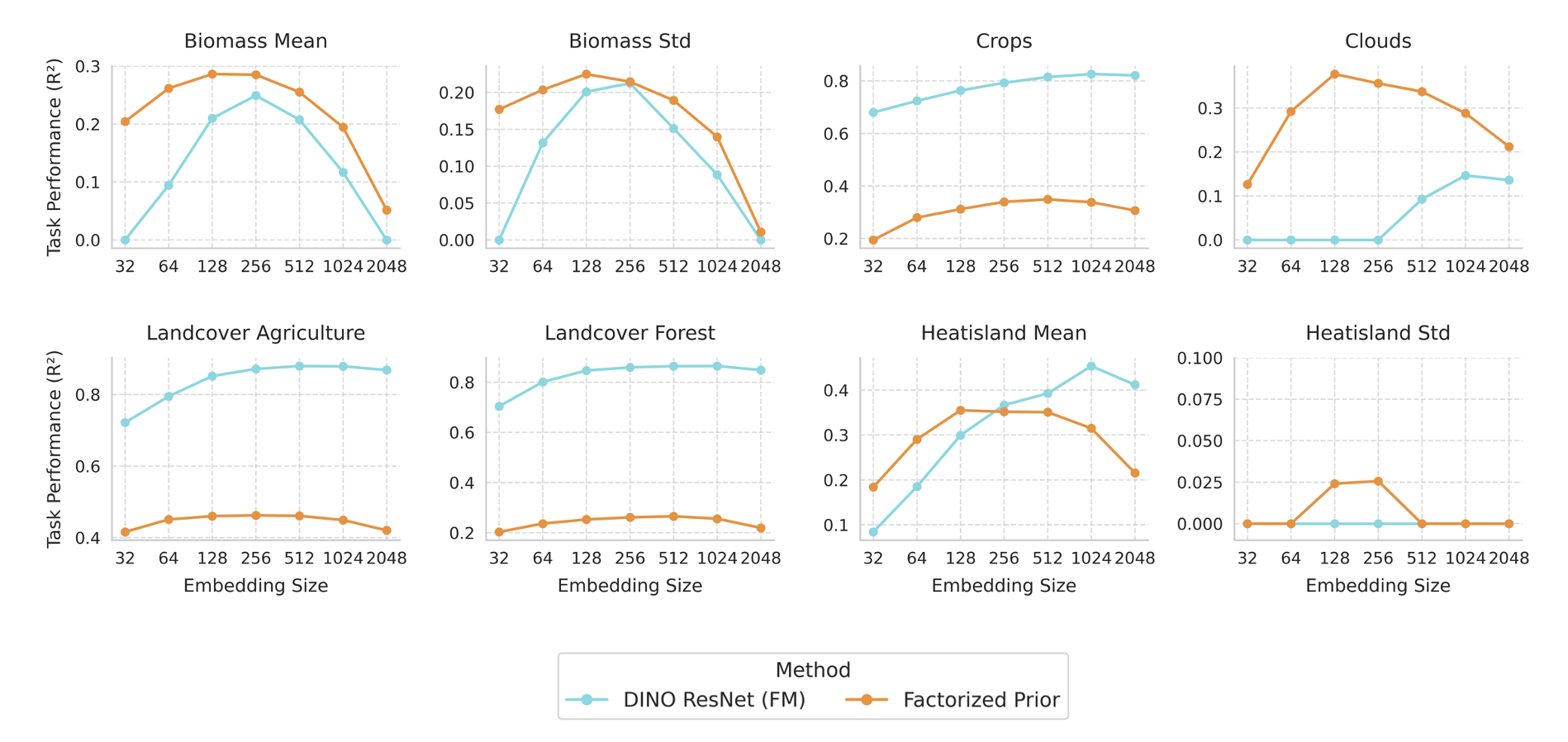}
  \caption{Embedding size ablation for CNN-based models. Optimal performance occurs between 128--1024 dimensions, with degradation outside this range.}
  \label{fig:size_cnn}
\end{figure*}

\paragraph{Embedding Size Ablations. }
Figures~\ref{fig:size_vit} and \ref{fig:size_cnn} show per-task ablation results on embedding dimensionality for ViT-based and CNN-based models, respectively. For CNN backbones, performance generally peaks in the range of 128–1024 dimensions, with larger or smaller embeddings leading to consistent performance drops. ViT-based embeddings, by contrast, are most effective at their natural patch-token dimension, and reductions tend to degrade task performance. Notably, the benefit of larger embeddings is limited: increases beyond 1024 dimensions yield negligible accuracy improvements while substantially raising computational demands and probe parameter counts. These results justify the use of 1024-dimensional embeddings as a balanced default in the run data challenge, while also illustrating \ifthenelse{\boolean{blind}}{our benchmark}{NeuCo-Bench}’s flexibility for exploring embedding-size vs. utility trade-offs in future studies.



%% file: main.bbl
\begin{thebibliography}{80}
\providecommand{\natexlab}[1]{#1}
\providecommand{\url}[1]{\texttt{#1}}
\expandafter\ifx\csname urlstyle\endcsname\relax
  \providecommand{\doi}[1]{doi: #1}\else
  \providecommand{\doi}{doi: \begingroup \urlstyle{rm}\Url}\fi

\bibitem[GEE()]{GEE}
{Google Earth Engine}.
\newblock \url{https://earthengine.google.com/}.
\newblock Accessed: 2025-05-14.

\bibitem[SSL(2025)]{SSL4EO-downstream}
{SSL4EO-S12-downstream}.
\newblock
  \url{https://huggingface.co/datasets/embed2scale/SSL4EO-S12-downstream},
  2025.
\newblock Accessed: 2025-05-22.

\bibitem[Alves~de Oliveira et~al.(2021)Alves~de Oliveira, Chabert, Oberlin,
  Poulliat, Bruno, Latry, Carlavan, Henrot, Falzon, and
  Camarero]{alves_de_oliveira_reduced-complexity_2021}
Vinicius Alves~de Oliveira, Marie Chabert, Thomas Oberlin, Charly Poulliat,
  Mickael Bruno, Christophe Latry, Mikael Carlavan, Simon Henrot, Frederic
  Falzon, and Roberto Camarero.
\newblock Reduced-complexity end-to-end variational autoencoder for on board
  satellite image compression.
\newblock \emph{Remote Sensing}, 2021.

\bibitem[Aybar et~al.(2024)Aybar, Bautista, Montero, Contreras, Ayala,
  Prudencio, Loja, Ysuhuaylas, Herrera, Gonzales, Valladares, Flores, Mamani,
  Quiñonez, Fajardo, Espinoza, Limas, Yali, Alcántara, Leyva, Loayza-Muro,
  Willems, Mateo-García, and Gómez-Chova]{Ayabar2024CloudSen12}
Cesar Aybar, Lesly Bautista, David Montero, Julio Contreras, Daryl Ayala,
  Fernando Prudencio, Jhomira Loja, Luis Ysuhuaylas, Fernando Herrera, Karen
  Gonzales, Jeanett Valladares, Lucy~A. Flores, Evelin Mamani, Maria Quiñonez,
  Rai Fajardo, Wendy Espinoza, Antonio Limas, Roy Yali, Alejandro Alcántara,
  Martin Leyva, Raúl Loayza-Muro, Bram Willems, Gonzalo Mateo-García, and
  Luis Gómez-Chova.
\newblock Cloudsen12+: The largest dataset of expert-labeled pixels for cloud
  and cloud shadow detection in sentinel-2.
\newblock \emph{Data in Brief}, 56:\penalty0 110852, 2024.
\newblock ISSN 2352-3409.
\newblock \doi{https://doi.org/10.1016/j.dib.2024.110852}.
\newblock URL
  \url{https://www.sciencedirect.com/science/article/pii/S2352340924008163}.

\bibitem[Ball\'{e} et~al.(2016)Ball\'{e}, Laparra, and
  Simoncelli]{balleEndtoendOptimizedImage2016}
Johannes Ball\'{e}, Valero Laparra, and Eero~P. Simoncelli.
\newblock End-to-end {{Optimized Image Compression}}.
\newblock \emph{International {{Conference}} on {{Learning Representations}}},
  2016.

\bibitem[Ball\'{e} et~al.(2018)Ball\'{e}, Minnen, Singh, Hwang, and
  Johnston]{balleVariationalImageCompression2018}
Johannes Ball\'{e}, David Minnen, Saurabh Singh, Sung~Jin Hwang, and Nick
  Johnston.
\newblock Variational image compression with a scale hyperprior.
\newblock In \emph{International {{Conference}} on {{Learning
  Representations}}}, 2018.

\bibitem[Bastani et~al.(2023)Bastani, Wolters, Gupta, Ferdinando, and
  Kembhavi]{satlas}
Favyen Bastani, Piper Wolters, Ritwik Gupta, Joe Ferdinando, and Aniruddha
  Kembhavi.
\newblock Satlaspretrain: A large-scale dataset for remote sensing image
  understanding, 2023.
\newblock URL \url{https://arxiv.org/abs/2211.15660}.

\bibitem[Blumenstiel et~al.(2025)Blumenstiel, Braham, Albrecht, Maurogiovanni,
  and Fraccaro]{blumenstiel2025ssl4eos12v11}
Benedikt Blumenstiel, Nassim Ait~Ali Braham, Conrad~M Albrecht, Stefano
  Maurogiovanni, and Paolo Fraccaro.
\newblock Ssl4eo-s12 v1.1: A multimodal, multiseasonal dataset for pretraining,
  updated, 2025.
\newblock URL \url{https://arxiv.org/abs/2503.00168}.

\bibitem[Boryan et~al.(2011)Boryan, Yang, Mueller, and Craig]{Boryan_CDL}
C.~Boryan, Z.~Yang, R.~Mueller, and M.~Craig.
\newblock Monitoring us agriculture: the us department of agriculture, national
  agricultural statistics service, cropland data layer program.
\newblock \emph{Geocarto International}, 26\penalty0 (5):\penalty0 341--358,
  2011.
\newblock \doi{10.1080/10106049.2011.562309}.
\newblock Dataset accessed via Google Earth Engine Data Catalog:
  \url{https://developers.google.com/earth-engine/datasets/catalog/USDA_NASS_CDL}
  (Accessed on 13.05.2025).

\bibitem[Bracewell(1986)]{fourier}
Ronald~Newbold Bracewell.
\newblock \emph{The Fourier transform and its applications}.
\newblock McGraw-Hill, 1986.

\bibitem[Brown et~al.(2025)Brown, Kazmierski, Pasquarella, Rucklidge,
  Samsikova, Zhang, Shelhamer, Lahera, Wiles, Ilyushchenko, Gorelick, Zhang,
  Alj, Schechter, Askay, Guinan, Moore, Boukouvalas, and
  Kohli]{brown2025alphaearth}
Christopher~F. Brown, Michal~R. Kazmierski, Valerie~J. Pasquarella, William~J.
  Rucklidge, Masha Samsikova, Chenhui Zhang, Evan Shelhamer, Estefania Lahera,
  Olivia Wiles, Simon Ilyushchenko, Noel Gorelick, Lihui~Lydia Zhang, Sophia
  Alj, Emily Schechter, Sean Askay, Oliver Guinan, Rebecca Moore, Alexis
  Boukouvalas, and Pushmeet Kohli.
\newblock Alphaearth foundations: An embedding field model for accurate and
  efficient global mapping from sparse label data, 2025.
\newblock URL \url{https://arxiv.org/abs/2507.22291}.

\bibitem[Cao et~al.(2022)Cao, Zhang, Zhao, Hu, and
  Wang]{cao_spectralspatial_2022}
T.~Cao, N.~Zhang, S.~Zhao, K.~Hu, and K.~Wang.
\newblock Spectral–spatial feature completely separated extraction with
  tensor {CNN} for multispectral image compression.
\newblock In \emph{Lecture Notes in Electrical Engineering}, 2022.

\bibitem[Chamain et~al.(2020)Chamain, Racapé, Bégaint, Pushparaja, and
  Feltman]{chamain2020endtoendoptimizedimagecompression}
Lahiru~D. Chamain, Fabien Racapé, Jean Bégaint, Akshay Pushparaja, and Simon
  Feltman.
\newblock End-to-end optimized image compression for machines, a study, 2020.
\newblock URL \url{https://arxiv.org/abs/2011.06409}.

\bibitem[Chamain et~al.(2021)Chamain, Racapé, Bégaint, Pushparaja, and
  Feltman]{chamain2021endtoendoptimizedimagecompression}
Lahiru~D. Chamain, Fabien Racapé, Jean Bégaint, Akshay Pushparaja, and Simon
  Feltman.
\newblock End-to-end optimized image compression for multiple machine tasks,
  2021.
\newblock URL \url{https://arxiv.org/abs/2103.04178}.

\bibitem[Codevilla et~al.(2021)Codevilla, Simard, Goroshin, and
  Pal]{codevilla2021learnedimagecompressionmachine}
Felipe Codevilla, Jean~Gabriel Simard, Ross Goroshin, and Chris Pal.
\newblock Learned image compression for machine perception, 2021.
\newblock URL \url{https://arxiv.org/abs/2111.02249}.

\bibitem[Daubechies(1992)]{wavelets}
Ingrid Daubechies.
\newblock \emph{Ten lectures on wavelets}.
\newblock Society for industrial and applied mathematics, USA, 1992.

\bibitem[Dosovitskiy et~al.(2020)Dosovitskiy, Beyer, Kolesnikov, Weissenborn,
  Zhai, Unterthiner, Dehghani, Minderer, Heigold, Gelly, Uszkoreit, and
  Houlsby]{vit}
Alexey Dosovitskiy, Lucas Beyer, Alexander Kolesnikov, Dirk Weissenborn,
  Xiaohua Zhai, Thomas Unterthiner, Mostafa Dehghani, Matthias Minderer, Georg
  Heigold, Sylvain Gelly, Jakob Uszkoreit, and Neil Houlsby.
\newblock An image is worth 16x16 words: Transformers for image recognition at
  scale.
\newblock \emph{CoRR}, abs/2010.11929, 2020.
\newblock URL \url{https://arxiv.org/abs/2010.11929}.

\bibitem[Du et~al.(2024)Du, Cheng, Olsen, Noghabi, Chandra, and
  Jiang]{du2024earth+}
Kuntai Du, Yihua Cheng, Peder Olsen, Shadi Noghabi, Ranveer Chandra, and
  Junchen Jiang.
\newblock Earth+: on-board satellite imagery compression leveraging historical
  earth observations.
\newblock \emph{arXiv preprint arXiv:2403.11434}, 2024.

\bibitem[Duan et~al.(2023)Duan, Ma, and Zhu]{compressed_domain_inference}
Zhihao Duan, Zhan Ma, and Fengqing Zhu.
\newblock Unified architecture adaptation for compressed domain semantic
  inference.
\newblock \emph{IEEE Transactions on Circuits and Systems for Video
  Technology}, 33\penalty0 (8):\penalty0 4108--4121, 2023.
\newblock \doi{10.1109/TCSVT.2023.3240391}.

\bibitem[Dubayah et~al.(2022)Dubayah, Armston, Kellner, Duncanson, Healey,
  Patterson, Hancock, Tang, Bruening, Hofton, Blair, and Luthcke]{GEDI}
R.O. Dubayah, J.~Armston, J.R. Kellner, L.~Duncanson, S.P. Healey, P.L.
  Patterson, S.~Hancock, H.~Tang, J.~Bruening, M.A. Hofton, J.B. Blair, and
  S.B. Luthcke.
\newblock Gedi l4a footprint level aboveground biomass density, version 2.1,
  2022.

\bibitem[Dubois et~al.(2022)Dubois, Bloem-Reddy, Ullrich, and
  Maddison]{dubois2022lossycompressionlosslessprediction}
Yann Dubois, Benjamin Bloem-Reddy, Karen Ullrich, and Chris~J. Maddison.
\newblock Lossy compression for lossless prediction, 2022.
\newblock URL \url{https://arxiv.org/abs/2106.10800}.

\bibitem[Dupont et~al.(2021)Dupont, Goli{\'n}ski, Alizadeh, Teh, and
  Doucet]{dupont2021coin}
Emilien Dupont, Adam Goli{\'n}ski, Milad Alizadeh, Yee~Whye Teh, and Arnaud
  Doucet.
\newblock {COIN}: Compression with implicit neural representations.
\newblock \emph{arXiv preprint arXiv:2103.03123}, 2021.

\bibitem[{European Environment Agency (EEA)}(2018)]{CLC2018}
{European Environment Agency (EEA)}.
\newblock Corine land cover (clc) 2018, version 20b, 100m raster.
\newblock
  \url{https://land.copernicus.eu/pan-european/corine-land-cover/clc2018},
  2018.
\newblock Accessed on 13.05.2025. Dataset accessed via Google Earth Engine
  dataset ID: COPERNICUS/CORINE/V20\_100m.

\bibitem[Fischer et~al.(2025)Fischer, Brand, and Kaup]{Fischer_2025}
Kristian Fischer, Fabian Brand, and André Kaup.
\newblock Boosting neural image compression for machines using latent space
  masking.
\newblock \emph{IEEE Transactions on Circuits and Systems for Video
  Technology}, 35\penalty0 (4):\penalty0 3719–3731, April 2025.
\newblock ISSN 1558-2205.
\newblock \doi{10.1109/tcsvt.2022.3195322}.
\newblock URL \url{http://dx.doi.org/10.1109/TCSVT.2022.3195322}.

\bibitem[Fuller et~al.(2023)Fuller, Millard, and Green]{croma}
Anthony Fuller, Koreen Millard, and James~R. Green.
\newblock Croma: Remote sensing representations with contrastive radar-optical
  masked autoencoders, 2023.
\newblock URL \url{https://arxiv.org/abs/2311.00566}.

\bibitem[Gomes and Brunschwiler(2024)]{Gomes2024NeuralEC}
Carlos Gomes and Thomas Brunschwiler.
\newblock Neural embedding compression for efficient multi-task earth
  observation modelling.
\newblock \emph{IGARSS 2024 - 2024 IEEE International Geoscience and Remote
  Sensing Symposium}, pages 8268--8273, 2024.
\newblock URL \url{https://api.semanticscholar.org/CorpusID:268692189}.

\bibitem[Gomes et~al.(2025)Gomes, Wittmann, Robert, Jakubik, Reichelt,
  Maurogiovanni, Vinge, Hurst, Scheurer, Sedona, et~al.]{gomes2025lossy}
Carlos Gomes, Isabelle Wittmann, Damien Robert, Johannes Jakubik, Tim Reichelt,
  Stefano Maurogiovanni, Rikard Vinge, Jonas Hurst, Erik Scheurer, Rocco
  Sedona, et~al.
\newblock Lossy neural compression for geospatial analytics: A review.
\newblock \emph{IEEE Geoscience and Remote Sensing Magazine}, 2025.

\bibitem[Gorelick et~al.(2017)Gorelick, Hancher, Dixon, Ilyushchenko, Thau, and
  Moore]{GORELICK201718}
Noel Gorelick, Matt Hancher, Mike Dixon, Simon Ilyushchenko, David Thau, and
  Rebecca Moore.
\newblock Google earth engine: Planetary-scale geospatial analysis for
  everyone.
\newblock \emph{Remote Sensing of Environment}, 202:\penalty0 18--27, 2017.
\newblock ISSN 0034-4257.
\newblock \doi{https://doi.org/10.1016/j.rse.2017.06.031}.
\newblock URL
  \url{https://www.sciencedirect.com/science/article/pii/S0034425717302900}.
\newblock Big Remotely Sensed Data: tools, applications and experiences.

\bibitem[Goyal(2001)]{goyal_theoretical_2001}
V.K. Goyal.
\newblock Theoretical foundations of transform coding.
\newblock \emph{IEEE Signal Processing Magazine}, 18\penalty0 (5):\penalty0
  9--21, September 2001.
\newblock ISSN 10535888.
\newblock \doi{10.1109/79.952802}.
\newblock URL \url{http://ieeexplore.ieee.org/document/952802/}.

\bibitem[Guo et~al.(2017)Guo, Liu, Jiang, Wang, Liu, and Liang]{guo_big_2017}
Huadong Guo, Zhen Liu, Hao Jiang, Changlin Wang, Jie Liu, and Dong Liang.
\newblock Big earth data: a new challenge and opportunity for digital earth’s
  development.
\newblock \emph{International Journal of Digital Earth}, 2017.

\bibitem[He et~al.(2015)He, Zhang, Ren, and Sun]{resnet}
Kaiming He, Xiangyu Zhang, Shaoqing Ren, and Jian Sun.
\newblock Deep residual learning for image recognition.
\newblock \emph{CoRR}, abs/1512.03385, 2015.
\newblock URL \url{http://arxiv.org/abs/1512.03385}.

\bibitem[Hong et~al.(2023)Hong, Zhang, Li, Li, Li, Yao, Yokoya, Li, Jia, Plaza,
  et~al.]{hong2023spectralgpt}
Danfeng Hong, Bing Zhang, Xuyang Li, Yuxuan Li, Chenyu Li, Jing Yao, Naoto
  Yokoya, Hao Li, Xiuping Jia, Antonio Plaza, et~al.
\newblock Spectralgpt: Spectral foundation model.
\newblock \emph{arXiv preprint arXiv:2311.07113}, 2023.

\bibitem[Huang and Wu(2024)]{huang_machinecoding}
Chen-Hsiu Huang and Ja-Ling Wu.
\newblock Unveiling the future of human and machine coding: A survey of
  end-to-end learned image compression.
\newblock \emph{Entropy}, 03 2024.
\newblock \doi{10.20944/preprints202403.1272.v1}.

\bibitem[Jakubik et~al.(2023)Jakubik, Roy, Phillips, Fraccaro, Godwin,
  Zadrozny, Szwarcman, Gomes, Nyirjesy, Edwards, et~al.]{jakubik2023foundation}
Johannes Jakubik, Sujit Roy, CE~Phillips, Paolo Fraccaro, Denys Godwin, Bianca
  Zadrozny, Daniela Szwarcman, Carlos Gomes, Gabby Nyirjesy, Blair Edwards,
  et~al.
\newblock Foundation models for generalist geospatial artificial intelligence.
\newblock \emph{arXiv preprint arXiv:2310.18660}, 2023.

\bibitem[Jakubik et~al.(2025)Jakubik, Yang, Blumenstiel, Scheurer, Sedona,
  Maurogiovanni, Bosmans, Dionelis, Marsocci, Kopp, Ramachandran, Fraccaro,
  Brunschwiler, Cavallaro, Bernabe-Moreno, and Longépé]{terramind}
Johannes Jakubik, Felix Yang, Benedikt Blumenstiel, Erik Scheurer, Rocco
  Sedona, Stefano Maurogiovanni, Jente Bosmans, Nikolaos Dionelis, Valerio
  Marsocci, Niklas Kopp, Rahul Ramachandran, Paolo Fraccaro, Thomas
  Brunschwiler, Gabriele Cavallaro, Juan Bernabe-Moreno, and Nicolas Longépé.
\newblock Terramind: Large-scale generative multimodality for earth
  observation, 2025.
\newblock URL \url{https://arxiv.org/abs/2504.11171}.

\bibitem[Klemmer et~al.(2025)Klemmer, Rolf, Robinson, Mackey, and
  Rußwurm]{klemmer2025satclip}
Konstantin Klemmer, Esther Rolf, Caleb Robinson, Lester Mackey, and Marc
  Rußwurm.
\newblock Satclip: Global, general-purpose location embeddings with satellite
  imagery.
\newblock \emph{Proceedings of the AAAI Conference on Artificial Intelligence},
  39\penalty0 (4):\penalty0 4347--4355, Apr. 2025.
\newblock \doi{10.1609/aaai.v39i4.32457}.

\bibitem[Kong et~al.(2021)Kong, Hu, Li, Li, and
  Zhao]{kong_spectralspatial_2021}
F.~Kong, K.~Hu, Y.~Li, D.~Li, and S.~Zhao.
\newblock Spectral–spatial feature partitioned extraction based on {CNN} for
  multispectral image compression.
\newblock \emph{Remote Sensing}, 2021.

\bibitem[Lacoste et~al.(2023)Lacoste, Lehmann, Rodriguez, Sherwin, Kerner,
  Lütjens, Irvin, Dao, Alemohammad, Drouin, Gunturkun, Huang, Vazquez, Newman,
  Bengio, Ermon, and Zhu]{lacoste2023geobenchfoundationmodelsearth}
Alexandre Lacoste, Nils Lehmann, Pau Rodriguez, Evan~David Sherwin, Hannah
  Kerner, Björn Lütjens, Jeremy~Andrew Irvin, David Dao, Hamed Alemohammad,
  Alexandre Drouin, Mehmet Gunturkun, Gabriel Huang, David Vazquez, Dava
  Newman, Yoshua Bengio, Stefano Ermon, and Xiao~Xiang Zhu.
\newblock Geo-bench: Toward foundation models for earth monitoring, 2023.
\newblock URL \url{https://arxiv.org/abs/2306.03831}.

\bibitem[Le et~al.(2021)Le, Zhang, Cricri, Ghaznavi-Youvalari, and
  Rahtu]{le2021imc}
Nam Le, Honglei Zhang, Francesco Cricri, Ramin Ghaznavi-Youvalari, and Esa
  Rahtu.
\newblock Image coding for machines: an end-to-end learned approach.
\newblock In \emph{ICASSP 2021 - 2021 IEEE International Conference on
  Acoustics, Speech and Signal Processing (ICASSP)}, pages 1590--1594, 2021.
\newblock \doi{10.1109/ICASSP39728.2021.9414465}.

\bibitem[Li et~al.(2022)Li, Hong, Gao, Yao, Zheng, Zhang, and
  Chanussot]{li2022deeplearningmultimodalremote}
Jiaxin Li, Danfeng Hong, Lianru Gao, Jing Yao, Ke~Zheng, Bing Zhang, and
  Jocelyn Chanussot.
\newblock Deep learning in multimodal remote sensing data fusion: A
  comprehensive review, 2022.
\newblock URL \url{https://arxiv.org/abs/2205.01380}.

\bibitem[Li et~al.(2023)Li, Sun, Liao, and Zhao]{li2023remote_}
Xin Li, Baile Sun, Jixiu Liao, and Xiaofei Zhao.
\newblock Remote sensing image compression method based on implicit neural
  representation.
\newblock In \emph{Proceedings of the International Conference on Computing and
  Pattern Recognition}, pages 432--439, 2023.

\bibitem[Liu et~al.(2024)Liu, Chen, Guan, Zhou, Zhu, Ye, Fu, and
  Zhou]{liu2024remoteclip}
Fan Liu, Delong Chen, Zhangqingyun Guan, Xiaocong Zhou, Jiale Zhu, Qiaolin Ye,
  Liyong Fu, and Jun Zhou.
\newblock {RemoteCLIP: A vision language foundation model for remote sensing}.
\newblock \emph{IEEE Transactions on Geoscience and Remote Sensing}, 2024.

\bibitem[Lu et~al.(2024)Lu, Guo, Zimmer-Dauphinee, Nieusma, Wang,
  VanValkenburgh, Wernke, and Huo]{lu2024ai_EOFMreview}
Siqi Lu, Junlin Guo, James~R Zimmer-Dauphinee, Jordan~M Nieusma, Xiao Wang,
  Parker VanValkenburgh, Steven~A Wernke, and Yuankai Huo.
\newblock Ai foundation models in remote sensing: A survey.
\newblock \emph{arXiv preprint arXiv:2408.03464}, 2024.

\bibitem[Mai et~al.(2022)Mai, Cundy, Choi, Hu, Lao, and Ermon]{Mai2022geoai}
Gengchen Mai, Chris Cundy, Kristy Choi, Yingjie Hu, Ni~Lao, and Stefano Ermon.
\newblock Towards a foundation model for geospatial artificial intelligence
  (vision paper).
\newblock In \emph{Proceedings of the 30th International Conference on Advances
  in Geographic Information Systems}. Association for Computing Machinery,
  2022.
\newblock ISBN 9781450395298.
\newblock \doi{10.1145/3557915.3561043}.
\newblock URL \url{https://doi.org/10.1145/3557915.3561043}.

\bibitem[Marsocci et~al.(2025)Marsocci, Jia, Bellier, Kerekes, Zeng, Hafner,
  Gerard, Brune, Yadav, Shibli, Fang, Ban, Vergauwen, Audebert, and
  Nascetti]{marsocci2025pangaeaglobalinclusivebenchmark}
Valerio Marsocci, Yuru Jia, Georges~Le Bellier, David Kerekes, Liang Zeng,
  Sebastian Hafner, Sebastian Gerard, Eric Brune, Ritu Yadav, Ali Shibli, Heng
  Fang, Yifang Ban, Maarten Vergauwen, Nicolas Audebert, and Andrea Nascetti.
\newblock Pangaea: A global and inclusive benchmark for geospatial foundation
  models, 2025.
\newblock URL \url{https://arxiv.org/abs/2412.04204}.

\bibitem[Minnen and Singh(2020)]{minnen2020channel}
David Minnen and Saurabh Singh.
\newblock Channel-wise autoregressive entropy models for learned image
  compression.
\newblock In \emph{IEEE International Conference on Image Processing}, 2020.

\bibitem[Minnen et~al.(2018)Minnen, Ball\'{e}, and
  Toderici]{autoregressive_hyperprior}
David Minnen, Johannes Ball\'{e}, and George~D Toderici.
\newblock Joint autoregressive and hierarchical priors for learned image
  compression.
\newblock In \emph{Advances in Neural Information Processing Systems}, 2018.

\bibitem[Observation and Center(2020)]{landsat8}
Earth~Resources Observation and Science~(EROS) Center.
\newblock Landsat 8-9 operational land imager / thermal infrared sensor
  level-2, collection 2.
\newblock \emph{U.S. Geological Survey}, 2020.
\newblock Dataset accessed via Google Earth Engine Data Catalog:
  \url{https://developers.google.com/earth-engine/datasets/catalog/LANDSAT_LC08_C02_T1_L2}
  (Accessed on 14.05.2025).

\bibitem[Plachouras et~al.(2025)Plachouras, Guinot, Fazekas, Quinton, Benetos,
  and Pauwels]{plachouras2025unifiedrepresentation}
Christos Plachouras, Julien Guinot, George Fazekas, Elio Quinton, Emmanouil
  Benetos, and Johan Pauwels.
\newblock Towards a unified representation evaluation framework beyond
  downstream tasks, 2025.
\newblock URL \url{https://arxiv.org/abs/2505.06224}.

\bibitem[Pouyanfar et~al.(2018)Pouyanfar, Yang, Chen, Shyu, and
  Iyengar]{multimedia_bigdata}
Samira Pouyanfar, Yimin Yang, Shu-Ching Chen, Mei-Ling Shyu, and S.~S. Iyengar.
\newblock Multimedia big data analytics: A survey.
\newblock \emph{ACM Comput. Surv.}, 51\penalty0 (1), January 2018.
\newblock ISSN 0360-0300.
\newblock \doi{10.1145/3150226}.
\newblock URL \url{https://doi.org/10.1145/3150226}.

\bibitem[Qian et~al.(2022)Qian, Lin, Sun, Tan, and Jin]{qian2022entroformer}
Yichen Qian, Ming Lin, Xiuyu Sun, Zhiyu Tan, and Rong Jin.
\newblock Entroformer: A transformer-based entropy model for learned image
  compression, 2022.
\newblock URL \url{https://arxiv.org/abs/2202.05492}.

\bibitem[Rezasoltani and Qureshi(2024)]{rezasoltani2024hyperspectral_}
Shima Rezasoltani and Faisal~Z Qureshi.
\newblock Hyperspectral image compression using sampling and implicit neural
  representations.
\newblock \emph{TGRS}, 63:\penalty0 10804213, 2024.

\bibitem[Richardson(2010)]{H.264}
Iain~E. Richardson.
\newblock \emph{The H.264 Advanced Video Compression Standard}.
\newblock John Wiley \& Sons, 2nd edition, 2010.

\bibitem[Rolf et~al.(2021)Rolf, Proctor, Carleton, Bolliger, Shankar, Ishihara,
  Recht, and Hsiang]{rolf2021mosaiks}
Esther Rolf, Jonathan Proctor, Tamma Carleton, Ian Bolliger, Vaishaal Shankar,
  Miyabi Ishihara, Benjamin Recht, and Solomon Hsiang.
\newblock A generalizable and accessible approach to machine learning with
  global satellite imagery.
\newblock \emph{Nature communications}, 12\penalty0 (1):\penalty0 4392, 2021.

\bibitem[Sialelli et~al.(2025)Sialelli, Peters, Wegner, and
  Schindler]{Sialelli2025}
Ghjulia Sialelli, Torben Peters, Jan~D. Wegner, and Konrad Schindler.
\newblock Agbd: A global-scale biomass dataset.
\newblock \emph{ISPRS Annals of the Photogrammetry, Remote Sensing and Spatial
  Information Sciences}, X-G-2025:\penalty0 829–838, July 2025.
\newblock ISSN 2194-9050.
\newblock \doi{10.5194/isprs-annals-x-g-2025-829-2025}.
\newblock URL \url{http://dx.doi.org/10.5194/isprs-annals-X-G-2025-829-2025}.

\bibitem[Singh et~al.(2020)Singh, Abu-El-Haija, Johnston, Ball{\'e},
  Shrivastava, and Toderici]{singh2020end}
Saurabh Singh, Sami Abu-El-Haija, Nick Johnston, Johannes Ball{\'e}, Abhinav
  Shrivastava, and George Toderici.
\newblock End-to-end learning of compressible features.
\newblock In \emph{2020 IEEE International Conference on Image Processing
  (ICIP)}, 2020.

\bibitem[Sitzmann et~al.(2020)Sitzmann, Martel, Bergman, Lindell, and
  Wetzstein]{sitzmann2020implicit}
Vincent Sitzmann, Julien Martel, Alexander Bergman, David Lindell, and Gordon
  Wetzstein.
\newblock Implicit neural representations with periodic activation functions.
\newblock \emph{NIPS}, 33:\penalty0 7462--7473, 2020.

\bibitem[Skodras et~al.(2001)Skodras, Christopoulos, and Ebrahimi]{jpeg2000}
A.~Skodras, C.~Christopoulos, and T.~Ebrahimi.
\newblock The jpeg 2000 still image compression standard.
\newblock \emph{IEEE Signal Processing Magazine}, 18\penalty0 (5):\penalty0
  36--58, 2001.
\newblock \doi{10.1109/79.952804}.

\bibitem[Str{\"u}mpler et~al.(2022)Str{\"u}mpler, Postels, Yang, Gool, and
  Tombari]{strumpler2022implicit}
Yannick Str{\"u}mpler, Janis Postels, Ren Yang, Luc~Van Gool, and Federico
  Tombari.
\newblock Implicit neural representations for image compression.
\newblock In \emph{ECCV}, pages 74--91. Springer, 2022.

\bibitem[Sullivan et~al.(2012)Sullivan, Ohm, Han, and Wiegand]{hevc}
Gary~J. Sullivan, Jens-Rainer Ohm, Woo-Jin Han, and Thomas Wiegand.
\newblock Overview of the high efficiency video coding (hevc) standard.
\newblock \emph{IEEE Transactions on Circuits and Systems for Video
  Technology}, 2012.

\bibitem[Sun et~al.(2022)Sun, Wang, Lu, Zhu, Lu, He, Li, Rong, Yang, Chang,
  et~al.]{sun2022ringmo}
Xian Sun, Peijin Wang, Wanxuan Lu, Zicong Zhu, Xiaonan Lu, Qibin He, Junxi Li,
  Xuee Rong, Zhujun Yang, Hao Chang, et~al.
\newblock {RingMo: A remote sensing foundation model with masked image
  modeling}.
\newblock \emph{IEEE Transactions on Geoscience and Remote Sensing}, 2022.

\bibitem[Szwarcman et~al.(2025)Szwarcman, Roy, Fraccaro, Þorsteinn
  Elí~Gíslason, Blumenstiel, Ghosal, de~Oliveira, de~Sousa~Almeida, Sedona,
  Kang, Chakraborty, Wang, Gomes, Kumar, Truong, Godwin, Lee, Hsu, Asanjan,
  Mujeci, Shidham, Keenan, Arevalo, Li, Alemohammad, Olofsson, Hain, Kennedy,
  Zadrozny, Bell, Cavallaro, Watson, Maskey, Ramachandran, and Moreno]{prithvi}
Daniela Szwarcman, Sujit Roy, Paolo Fraccaro, Þorsteinn Elí~Gíslason,
  Benedikt Blumenstiel, Rinki Ghosal, Pedro~Henrique de~Oliveira, Joao~Lucas
  de~Sousa~Almeida, Rocco Sedona, Yanghui Kang, Srija Chakraborty, Sizhe Wang,
  Carlos Gomes, Ankur Kumar, Myscon Truong, Denys Godwin, Hyunho Lee, Chia-Yu
  Hsu, Ata~Akbari Asanjan, Besart Mujeci, Disha Shidham, Trevor Keenan, Paulo
  Arevalo, Wenwen Li, Hamed Alemohammad, Pontus Olofsson, Christopher Hain,
  Robert Kennedy, Bianca Zadrozny, David Bell, Gabriele Cavallaro, Campbell
  Watson, Manil Maskey, Rahul Ramachandran, and Juan~Bernabe Moreno.
\newblock Prithvi-eo-2.0: A versatile multi-temporal foundation model for earth
  observation applications, 2025.
\newblock URL \url{https://arxiv.org/abs/2412.02732}.

\bibitem[Theis et~al.(2022)Theis, Shi, Cunningham, and
  Husz{\'a}r]{theis2022lossy}
Lucas Theis, Wenzhe Shi, Andrew Cunningham, and Ferenc Husz{\'a}r.
\newblock Lossy image compression with compressive autoencoders.
\newblock In \emph{International conference on learning representations}, 2022.

\bibitem[Torfason et~al.(2018)Torfason, Mentzer, Agustsson, Tschannen, Timofte,
  and Van~Gool]{torfason2018imageunderstandingdeepcompression}
Robert Torfason, Fabian Mentzer, Eirikur Agustsson, Michael Tschannen, Radu
  Timofte, and Luc Van~Gool.
\newblock Towards image understanding from deep compression without decoding.
\newblock \emph{ArXiv}, abs/1803.06131, 2018.
\newblock URL \url{https://arxiv.org/abs/1803.06131}.

\bibitem[Wallace(1991)]{jpeg}
Gregory~K. Wallace.
\newblock The jpeg still picture compression standard.
\newblock \emph{Commun. ACM}, 1991.

\bibitem[Wang et~al.(2022{\natexlab{a}})Wang, Zhang, Xu, Zhang, Du, Tao, and
  Zhang]{wang2022advancing}
Di~Wang, Qiming Zhang, Yufei Xu, Jing Zhang, Bo~Du, Dacheng Tao, and Liangpei
  Zhang.
\newblock Advancing plain vision transformer toward remote sensing foundation
  model.
\newblock \emph{IEEE Transactions on Geoscience and Remote Sensing},
  2022{\natexlab{a}}.

\bibitem[Wang et~al.(2021)Wang, Wang, Wang, and Ye]{wang2021endtoend}
Shurun Wang, Zhao Wang, Shiqi Wang, and Yan Ye.
\newblock End-to-end compression towards machine vision: Network architecture
  design and optimization.
\newblock \emph{IEEE Open Journal of Circuits and Systems}, 2:\penalty0
  675--685, 2021.
\newblock \doi{10.1109/OJCAS.2021.3126061}.

\bibitem[Wang et~al.(2023{\natexlab{a}})Wang, Wang, Wang, and
  Ye]{wang2023deepimagecompression}
Shurun Wang, Zhao Wang, Shiqi Wang, and Yan Ye.
\newblock Deep image compression toward machine vision: A unified optimization
  framework.
\newblock \emph{IEEE Transactions on Circuits and Systems for Video
  Technology}, 33\penalty0 (6):\penalty0 2979--2989, 2023{\natexlab{a}}.
\newblock \doi{10.1109/TCSVT.2022.3230843}.

\bibitem[Wang et~al.(2018{\natexlab{a}})Wang, Hu, Wang, and
  Xiao]{video_earth_observation}
Xu~Wang, Ruimin Hu, Zhongyuan Wang, and Jing Xiao.
\newblock Virtual background reference frame based satellite video coding.
\newblock \emph{IEEE Signal Processing Letters}, 2018{\natexlab{a}}.

\bibitem[Wang et~al.(2022{\natexlab{b}})Wang, Albrecht, Braham, Mou, and
  Zhu]{wang2022selfsupervisedlearningremotesensing}
Yi~Wang, Conrad~M Albrecht, Nassim Ait~Ali Braham, Lichao Mou, and Xiao~Xiang
  Zhu.
\newblock Self-supervised learning in remote sensing: A review,
  2022{\natexlab{b}}.
\newblock URL \url{https://arxiv.org/abs/2206.13188}.

\bibitem[Wang et~al.(2023{\natexlab{b}})Wang, Braham, Xiong, Liu, Albrecht, and
  Zhu]{Wang23}
Yi~Wang, Nassim Ait~Ali Braham, Zhitong Xiong, Chenying Liu, Conrad~M.
  Albrecht, and Xiao~Xiang Zhu.
\newblock Ssl4eo-s12: A large-scale multimodal, multitemporal dataset for
  self-supervised learning in earth observation [software and data sets].
\newblock \emph{IEEE Geoscience and Remote Sensing Magazine}, 11\penalty0
  (3):\penalty0 98--106, 2023{\natexlab{b}}.
\newblock \doi{10.1109/MGRS.2023.3281651}.

\bibitem[Wang et~al.(2025)Wang, Xiong, Liu, Stewart, Dujardin, Bountos, Zavras,
  Gerken, Papoutsis, Leal-Taixé, and Zhu]{Wang_2025_ICCV_CopernicusFM}
Yi~Wang, Zhitong Xiong, Chenying Liu, Adam~J. Stewart, Thomas Dujardin,
  Nikolaos~Ioannis Bountos, Angelos Zavras, Franziska Gerken, Ioannis
  Papoutsis, Laura Leal-Taixé, and Xiao~Xiang Zhu.
\newblock Towards a unified copernicus foundation model for earth vision.
\newblock In \emph{Proceedings of the IEEE/CVF International Conference on
  Computer Vision (ICCV)}, pages 9888--9899, October 2025.

\bibitem[Wang et~al.(2018{\natexlab{b}})Wang, Mao, Yang, and
  Tang]{survey_multimedia}
Zaijian Wang, Shiwen Mao, Lingyun Yang, and Pingping Tang.
\newblock A survey of multimedia big data.
\newblock \emph{China Communications}, 15\penalty0 (1):\penalty0 155--176,
  2018{\natexlab{b}}.
\newblock \doi{10.1109/CC.2018.8290814}.

\bibitem[Wiesenfarth et~al.(2021)Wiesenfarth, Reinke, Landman, Eisenmann, Saiz,
  Cardoso, Maier-Hein, and Kopp-Schneider]{wiesenfarth2021methods}
Manuel Wiesenfarth, Annika Reinke, Bennett~A Landman, Matthias Eisenmann,
  Laura~Aguilera Saiz, M~Jorge Cardoso, Lena Maier-Hein, and Annette
  Kopp-Schneider.
\newblock Methods and open-source toolkit for analyzing and visualizing
  challenge results.
\newblock \emph{Scientific reports}, 11\penalty0 (1):\penalty0 2369, 2021.

\bibitem[Wilkinson et~al.(2024)Wilkinson, Mleczko, Brewin, Gaston, Mueller,
  Shutler, Yan, and Anderson]{wilkinson_environmental_2024}
R.~Wilkinson, M.~M. Mleczko, R.~J.~W. Brewin, K.~J. Gaston, M.~Mueller, J.~D.
  Shutler, X.~Yan, and K.~Anderson.
\newblock Environmental impacts of earth observation data in the constellation
  and cloud computing era.
\newblock \emph{Science of The Total Environment}, 2024.

\bibitem[Xiong et~al.(2024)Xiong, Wang, Zhang, Stewart, Hanna, Borth,
  Papoutsis, Saux, Camps-Valls, and Zhu]{dofa}
Zhitong Xiong, Yi~Wang, Fahong Zhang, Adam~J. Stewart, Joëlle Hanna, Damian
  Borth, Ioannis Papoutsis, Bertrand~Le Saux, Gustau Camps-Valls, and
  Xiao~Xiang Zhu.
\newblock Neural plasticity-inspired multimodal foundation model for earth
  observation, 2024.
\newblock URL \url{https://arxiv.org/abs/2403.15356}.

\bibitem[Xu and Tewari(2021)]{Ziping2021}
Ziping Xu and Ambuj Tewari.
\newblock Representation learning beyond linear prediction functions.
\newblock In M.~Ranzato, A.~Beygelzimer, Y.~Dauphin, P.S. Liang, and J.~Wortman
  Vaughan, editors, \emph{Advances in Neural Information Processing Systems},
  volume~34, pages 4792--4804. Curran Associates, Inc., 2021.
\newblock URL
  \url{https://proceedings.neurips.cc/paper_files/paper/2021/file/258be18e31c8188555c2ff05b4d542c3-Paper.pdf}.

\bibitem[Yadav et~al.()Yadav, Jain, Agrawal, and
  Chattopadhyay]{yadavEvalAIBetterEvaluation}
Deshraj Yadav, Rishabh Jain, Harsh Agrawal, and Prithvijit Chattopadhyay.
\newblock {{EvalAI}}: {{Towards Better Evaluation}} of {{AI Agents}}.
\newblock \emph{EvalAI}.
\newblock URL \url{https://eval.ai/}.

\bibitem[Yeh et~al.(2005)Yeh, Armbruster, Kiely, Masschelein, Moury, Schaefer,
  and Thiebaut]{yeh_new_2005}
Pen-Shu Yeh, P.~Armbruster, A.~Kiely, B.~Masschelein, G.~Moury, C.~Schaefer,
  and C.~Thiebaut.
\newblock The new {CCSDS} image compression recommendation.
\newblock In \emph{{IEEE} Aerospace Conference}, 2005.

\bibitem[Zhang et~al.(2024)Zhang, Pan, Liu, and Han]{zhang2024compressing_}
Lili Zhang, Tianpeng Pan, Jiahui Liu, and Lin Han.
\newblock Compressing hyperspectral images into multilayer perceptrons using
  fast-time hyperspectral neural radiance fields.
\newblock \emph{GRSL}, 21:\penalty0 10433191, 2024.

\end{thebibliography}
